\documentclass[10pt,twocolumn,letterpaper]{article}

\usepackage{iccv}              
\usepackage{amsmath}
\usepackage{xparse}
\usepackage{xspace}
\usepackage{diagbox}        %
\usepackage{booktabs}       
\usepackage{multirow}
\usepackage{threeparttable}
\usepackage{array}
\usepackage{colortbl}
\usepackage{times}
\usepackage{epsfig}
\usepackage{graphicx}
\usepackage{enumitem}
\usepackage{pifont}
\usepackage{amsfonts}
\usepackage{amssymb,amsthm}
\usepackage{arydshln}
\usepackage{float}
\usepackage[pagebackref,breaklinks,colorlinks]{hyperref}
\usepackage{balance}
\usepackage{tipa}
\usepackage{picins}
\usepackage{fontawesome5}
\usepackage{color}
\usepackage{tikz}
\usepackage{microtype}
\usepackage{stfloats}
\usepackage{bigstrut,rotating}
\usepackage[utf8]{inputenc}
\usepackage[T1]{fontenc}
\usepackage{bm}
\usepackage{siunitx}
\usepackage[nopar]{lipsum}
\usepackage[export]{adjustbox}

\usepackage[lined,boxed,commentsnumbered,ruled]{algorithm2e}
\usetikzlibrary{arrows,shadows} 
\usepackage[underline=true,rounded corners=false]{pgf-umlsd}
\def\eg{\emph{e.g.}} 
\def\ie{\emph{i.e.}} 
 
\definecolor{bblue}{rgb}{0,150,230}
\definecolor{mygray}{gray}{.9}
\definecolor{lightgray}{gray}{.96}
\definecolor{myy}{RGB}{126,95,0}
\definecolor{ggray}{RGB}{127,127,127}
\definecolor{mygreen}{RGB}{93,173,85}
\definecolor{myred}{RGB}{240,16,89}
\definecolor{myblue}{RGB}{0,114,188}
\definecolor{darkgreen}{rgb}{0.0, 0.5, 0.0}
\definecolor{demphcolor}{RGB}{100,100,100}

\definecolor{cvprblue}{rgb}{0.21,0.49,0.74}

\definecolor{darkcyan}{rgb}{0,.79,1} 
\definecolor{LightCyan}{rgb}{0.88,1,1} 
\definecolor{darkred}{rgb}{0.39,0.04,0.45} 
\definecolor{earthyellow1}{rgb}{1, 0.98, 0.98}
\definecolor{darkblue}{rgb}{0.34, 0.66, 0.70}
\definecolor{darkblue1}{rgb}{0.18, 0.30, 0.28}
\definecolor{Lightpurple}{rgb}{0.89, 0.83, 0.90}
\definecolor{Lightpurple1}{rgb}{1, 0, 0.55}
\definecolor{shadowred}{rgb}{0.92, 0.70, 0.79}
\definecolor{shadowblue}{rgb}{0.92, 0.97, 1}

\definecolor{shadowred}{rgb}{0.97, 0.68, 0.60}
\definecolor{shadowgreen}{rgb}{0.67, 0.97, 0.75}
\definecolor{lightblue}{rgb}{0.72, 0.75, 0.97}
\definecolor{lightgray}{rgb}{0.3, 0.85, 0.86}

\definecolor{highblue}{rgb}{0, 0.54, 0.9}
\definecolor{purple}{rgb}{1, 0.32, 1} 

\def\datasetname{Drive-Gaze}
\def\modelname{Causal-VidSyn}

\def\paperID{6974} 
\def\confName{ICCV}
\def\confYear{2025}

 \title {Causal-Entity Reflected Egocentric Traffic Accident Video Synthesis}

\author{Lei-lei Li$^{1}$, Jianwu Fang$^{1\dag}$, Junbin Xiao$^{2}$, Shanmin Pang$^{1}$, Hongkai Yu$^{4}$, Chen Lv$^{3}$, \\Jianru Xue$^{1}$, and Tat-Seng Chua$^{2}$ \\
\small{$^1$Xi'an Jiaotong University\quad$^2$National University of Singapore\quad$^3$Nanyang Technological University\quad$^4$Cleveland State University}\\
\small{\url{http://lotvsmmau.net/Causal-VidSyn}}
\thanks{\dag Corresponding author.}}

\begin{document}
\maketitle

\begin{abstract}
Egocentricly comprehending the causes and effects of car accidents is crucial for the safety of self-driving cars, and synthesizing causal-entity reflected accident videos can facilitate the capability test to respond to unaffordable accidents in reality.
However, incorporating causal relations as seen in real-world videos into synthetic videos remains challenging. This work argues that precisely identifying the accident participants and capturing their related behaviors are of critical importance. In this regard, we propose a novel diffusion model \modelname ~for synthesizing egocentric traffic accident videos. To enable causal entity grounding in video diffusion, \modelname ~leverages the cause descriptions and driver fixations to identify the accident participants and behaviors, facilitated by accident reason answering and gaze-conditioned selection modules.
To support \modelname, we further construct \textbf{\datasetname}, the largest driver gaze dataset (with \textbf{1.54M} frames of fixations) in driving accident scenarios. Extensive experiments show that \modelname ~surpasses state-of-the-art video diffusion models in terms of frame quality and causal sensitivity in various tasks, including accident video editing, normal-to-accident video diffusion, and text-to-video generation.
\end{abstract}
\vspace{-1.5em}

\section{Introduction}
The emerging fully self-driving (FSD) technique~\cite{FSD2024}, while bringing convenience to our daily life, has also led to various ethical, trustworthy, and economic disputes in handling car accident~\cite{DBLP:journals/natmi/X22a}. Therefore, an egocentric comprehension of the car accident is paramount not only for improving self-driving safety but also for disambiguating accident responsibility. Yet, the scarcity of egocentric accident data severely hinders the research in this field. 

\begin{figure}[!t]
    \centering    
    \includegraphics[width=\linewidth]{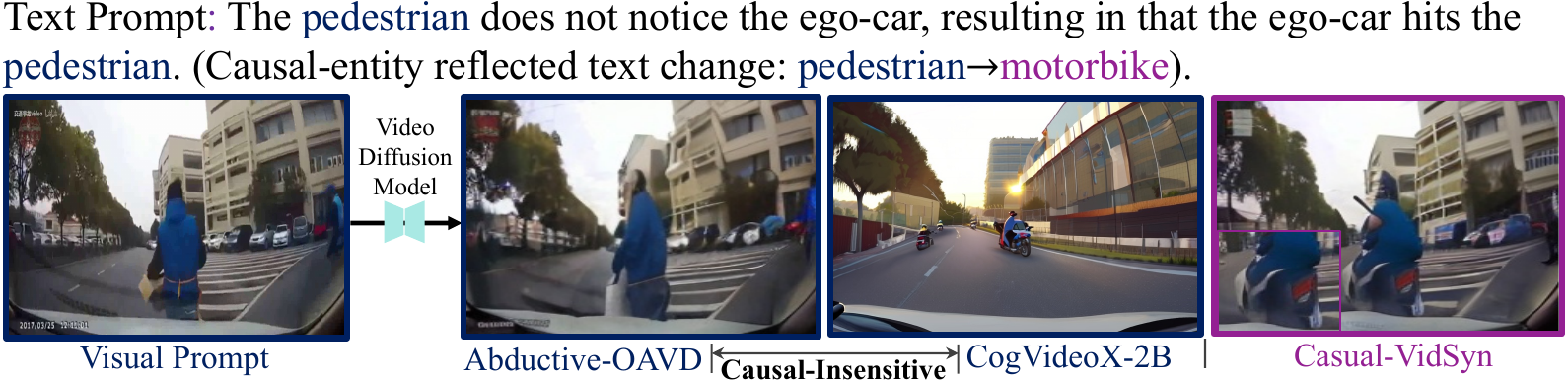}
        \vspace{-1.5em}
    \caption{We show the inability of two state-of-the-art diffusion models (\ie, \textcolor{myblue}{Abductive-OAVD}~\cite{DBLP:journals/corr/abs-2212-09381} and \textcolor{myblue}{CogVideoX-2B}~\cite{yang2024cogvideox}) for editing accident video content. Abductive-OAVD cannot generate the needed motorbike, while CogVideoX fails to reflect the collision situation. Our \textcolor{darkred}{\modelname }~accurately generates the collision motorbike and maintains the background scenes. }
    \label{fig1}
    \vspace{-1.5em}
\end{figure}

With the tremendous advancements in video diffusion models~\cite{DBLP:conf/iccv/KhachatryanMTHW23,DBLP:journals/corr/abs-2204-03458,song2020denoising,DBLP:conf/iccv/EsserCAGG23,DBLP:conf/aaai/YeB24,DBLP:conf/aaai/ChuHLC24,yang2025videograin}, synthetic videos could be a promising solution for data scarcity.
However, the state-of-the-art (SOTA) video diffusion models are developed for common video generation; they often fail to generate egocentric accident videos that achieve a causal-entity reflected  synthesis~\cite{DBLP:journals/corr/abs-2401-03048,yang2024cogvideox}. As shown in Fig.~\ref{fig1}, when facing a counterfactual text edit from ``\textcolor{myblue}{pedestrian}'' to ``\textcolor{darkred}{motorbike}'' collision, one of the SOTA model Abductive-OAVD~\cite{DBLP:journals/corr/abs-2212-09381} fails to generate the motorbike. While CogVideoX~\cite{yang2024cogvideox} responds to the needed motorbike, the requested collision is not reflected.

To overcome such issues, precisely identifying the causal entities (or objects) in accidents and capturing their accident-related behaviors is the
key to success. 
However, accident scenarios often involve tiny objects in fast scene changes, which makes it extremely difficult to identify the objects, especially in the ego view, not to mention analyzing their nuanced behaviors related to the accidents. In this paper, we highlight the incorporation of two critical information cues: accident reason-collision descriptions and driver gaze fixations. The reason-collision descriptions contain rich information about the accidents, including the major participants and their misbehavior that resulted in the accidents. Despite being informative, finding the right reason for the accident is still challenging. Also, there is a modality gap between textual descriptions to visual accident appearances.   
As a remedy, driver gaze fixations provide direct visual attention to the accident regions, since human drivers can perceive the road hazard incisively based on their driving experience.

To effectively exploit such information for egocentric accident video diffusion, 
we propose the \modelname, which makes the video diffusion backbone (\eg, 3D-Unet) causal-grounded, facilitated by 1) an accident reason answering (ArA) module designed to retrieve the right accident reason and incorporate it into noise representation learning, and 2) a driver gaze-conditioned visual token selection mechanism to focus on causal participant regions. Additionally, to enhance diffusion learning in fast scene changes, we additionally learn to contrast forward and backward time order frame diffusion with reciprocal text and vision prompts, as creating a counterfactual intervention to forward text and visual prompts, via the exogenous noise $\bm{e}$ to help the causal scene learning associating reciprocal frame and text prompts.

\modelname~fulfills the causal-entity reflected video synthesis from 
\ding{182} internal diffusion recipe level and \ding{183} external knowledge level. Diffusion recipe level fulfills a reciprocal prompted frame diffusion (RPFD, \S\ref{recipe-level}). The knowledge level designs \emph{causal-prone}\footnote{Causal-prone tokens mean that the selected tokens prefer better causal-sensitive text-vision alignment than non-selected ones.} token selection blocks (\textcolor{darkred}{\textbf{CTS}}) and an ArA-guided causal token grounding block (\textcolor{darkblue1}{\textbf{CTG}}) (\S\ref{model-level}) to make the 3D-Unet backbone causal-grounded.
 Additionally, to support \modelname, we construct the largest driver gaze dataset \textbf{\datasetname}, which collects over \textbf{$1.54$ million} frames of driver fixations for $9,727$ accident scenarios. Notably, ArA and driver fixations are only involved in the training phase, and the testing phase only inputs the video or text prompts for accident video diffusion.

Extensive evaluations are carried out for three video diffusion tasks 1) accident video content editing on accident dataset (\textbf{DADA-2000}~\cite{DBLP:journals/tits/FangYQXY22}), 2) normal-to-accident video diffusion on accident-free dataset (\textbf{BDD-A}~\cite{DBLP:conf/accv/XiaZKNZW18}), and 3) text-to-video accident generation. The results show that \modelname~can surpass many state-of-the-art methods with higher frame quality, fidelity, and causal sensitivity. We also extend \textcolor{darkred}{\textbf{CTS}} and \textcolor{darkblue1}{\textbf{CTG}} to SOTA Transformer-based video diffusion models (\emph{i.e.}, CogVideoX-2B~\cite{yang2024cogvideox} and Latte~\cite{DBLP:journals/corr/abs-2401-03048}), and demonstrate consistently significant improvements.

\section{Related Work}
\label{sec:related_work}
\subsection{Approaches for Ameliorating Diffusion Models}
\textbf{Enhancing Content Consistency}.
Video diffusion models have dominated the field of video generation in recent years~\cite{2detail,3align,16videofusion, 18lavie,DBLP:journals/corr/abs-2405-14864,hu2025videoshield,wang2025seedvrseedin} and take large efforts to overcome a knotty problem, \emph{i.e.}, maintaining temporal logic and content consistency~\cite{1storydiffusion, 3align, 4inflation}, such as StoryDiffusion~\cite{1storydiffusion}, CogVideoX~\cite{yang2024cogvideox}, T2V-Turbo-v2~\cite{li2024t2v}, etc. Various temporal alignment modules are modeled to ensure consistency across video frames in~\cite{DBLP:journals/visintelligence/RajendranTANS24,3align, 4inflation, 5towards, 8genvideo,35versvideo, 36learning}. \\
\textbf{Causality Discovery}.
Causal relations are natural in the real world~\cite{DBLP:journals/pami/WangWZFHC23,DBLP:conf/iclr/0003LMC0C23,DBLP:journals/pami/YangZC23,liu2024vdg,DBLP:conf/aaai/LiFZJW0ZL24}. Based on the discussion on causality and grounding~\cite{zhang2023unify}, causality prefers a time-different grounding between the cause and effect elements. Recent causal diffusion models mainly focus on the counterfactual (what-if issue) estimation~\cite{DBLP:conf/clear2/SanchezT22,DBLP:journals/corr/abs-2404-17735} or contrastive causal-relation discovery~\cite{DBLP:conf/iclr/SanchezLOT23,mamaghan2023diffusion}, \emph{e.g.}, forward noising the causal relations of latent variables step-by-step~\cite{DBLP:conf/clear2/SanchezT22,DBLP:conf/aaai/ManDN24}. 
Compared to the domain-extraneous ways, considering driving scene knowledge is preferred in this work. \\
\textbf{Considering Driving Scene Knowledge}.
Video diffusion in driving scenes has garnered widespread attention~\cite{ 14panacea, 20magicdrive, 32gaia,DBLP:journals/corr/abs-2411-15139,DBLP:journals/corr/abs-2412-13188} for safe driving by frame or view synthesis. The multi-view panoramic consistency within captured videos, such as in DriverDreamer~\cite{DBLP:journals/corr/abs-2309-09777}, DrivingDiffusion~\cite{13drivingdiffusion}, Panacea~\cite{14panacea}, \emph{etc.}, involve the road topology (\emph{e.g.}, bird's eye view (BEV)~\cite{15generating, 19stealing, 38text2street}) and object locations to correlate the ego-scene relation in multi-view driving video diffusion. Nonetheless, current video diffusion models mainly encounter accident-free scenarios, while the knowledge of on-road accident videos is commonly absent. Recent works explore the accident video synthesis~\cite{DBLP:journals/corr/abs-2212-09381,guo2024drivinggen,li2025avd2}, while they mainly focus on the text-to-video generation, and do not explore the causality in video conditioned synthesis. 

\subsection{Driver Gaze Datasets for Driving Videos}
Tab.~\ref {tab1} presents the attribute analysis for available driver gaze datasets. Among them, BDD-A~\cite{DBLP:conf/accv/XiaZKNZW18}, DADA-2000 \cite{DBLP:conf/itsc/FangYQXWL19}, and DR(eye)VE \cite{DBLP:journals/pami/PalazziACSC19} are the top-3 most popular ones for critical, accident, and normal driving scenarios, respectively. DR(eye)VE \cite{DBLP:journals/pami/PalazziACSC19}, LBW \cite{DBLP:conf/eccv/KasaharaSP22}, and DPoG \cite{DBLP:conf/eccv/KasaharaSP22} collect the gaze data in the real driving process. Most datasets focus on normal driving scenes. DADA-2000 \cite{DBLP:conf/itsc/FangYQXWL19}, Eye-car \cite{DBLP:conf/iccv/BaeePK0OB21}, PSAD \cite{DBLP:conf/itsc/GanLWCQN21}, and our \datasetname~concentrate on the driving accident scenarios.  \datasetname~is the largest driver gaze dataset and owns the text description for accident reasons and collision descriptions. 
\begin{table}[!t]\scriptsize
  \centering
  \caption{Attribute comparison of available driver gaze datasets.}
    \vspace{-1em}
 \setlength{\tabcolsep}{0.2mm}{
\begin{tabular}{l|ccccccccc}
 \toprule
Datasets &Years&\#Clips &\#Frames &CoT-F&S-C&Sub.num&TA&R/S&Cites\\
\hline
3DDS \cite{DBLP:conf/bmvc/BorjiSI11}  &2011& -&18K& in-lab &NOM &10&&S&56\\
BDD-A  \cite{DBLP:conf/accv/XiaZKNZW18} &2018& 1,232& 378K& in-lab&CR &45 &&R&181\\
DADA-2000 \cite{DBLP:conf/itsc/FangYQXWL19,DBLP:journals/tits/FangYQXY22}&2019&2,000&658K&in-lab&Acc&20&&R&244\\
DR(eye)VE \cite{DBLP:journals/pami/PalazziACSC19}&2019& 74& 555K&real-d & NOM& 8&&R&342\\
DGAZE \cite{DBLP:conf/iros/DuaJGJ20} &2020&20&100K& in-lab & NOM & 20& &R&21\\
TrafficGaze \cite{DBLP:journals/tits/DengYQNM20}  &2020& 16&75K& in-lab&NOM & 28 &&R&117\\
MAAD \cite{gopinath2021maad} &2021& 8&60K&in-lab &NOM& 23 &&R&13\\
Eye-car \cite{DBLP:conf/iccv/BaeePK0OB21} &2021&21&31.5K& in-lab& Acc & 20& &R&60\\
PSAD \cite{DBLP:conf/itsc/GanLWCQN21}  &2021&2,724&797K& in-lab&Acc & 6& &R&10\\
RainyGaze \cite{DBLP:journals/ieeejas/TianDY22}  &2022& 16&81K& in-lab& NOM&  30& &R&16\\
CoCAtt \cite{DBLP:conf/itsc/ShenWSHDC22}  &2022& -&88K& in-lab &NOM &  11& &S&13\\
LBW \cite{DBLP:conf/eccv/KasaharaSP22} &2022&-&123K& real-d&NOM& 28  & &R&27\\
DPoG \cite{nguyen2024driver} &2024&-&15.3K& real-d&NOM& 11  & &R&1\\
\hline
\textcolor{darkred}{\datasetname}  &\textcolor{darkred}{2025}&\textcolor{darkred}{\textbf{9,727}}&\textcolor{darkred}{\textbf{1.54 million}}& \textcolor{darkred}{in-lab} & \textcolor{darkred}{Acc}&\textcolor{darkred}{10} &\textcolor{darkred}{\checkmark} & \textcolor{darkred}{R}&\textcolor{darkred}{-}\\
    \hline
  \end{tabular}}
    \begin{tablenotes}
\item \scriptsize{\textbf{CoT-F}: collection form (in-lab or real driving (real-d)), \textbf{S-C}: scene categories (NOM-normal, CR-critical, and Acc-accident), \textbf{TA}: text annotations, \textbf{Sub.num}: subject number, \textbf{ReS}: frame resolution, \textbf{R/S}: real or synthetic videos. \textbf{Cites}: google citations up to Mar. 07, 2025.}
\end{tablenotes}
  \label{tab1}
  \vspace{-2em}
  \end{table}

 \section{\datasetname~Dataset}
The data source of Drive-Gaze stems from the recently released multimodal egocentric accident video dataset, \emph{i.e.}, MM-AU~\cite{DBLP:journals/corr/abs-2212-09381}, which annotates the accident reason, collision type, and accident prevention descriptions for 11,727 accident videos. We find that DADA-2000~\cite{DBLP:conf/itsc/FangYQXWL19} is MM-AU's subset, having the driver fixation data already. We collect the diver gaze data for the remainder accident videos. We have ten subjects (4 females and 6 males aged from 21 to 26 years old with over two years of driver's licenses) to collect the driver fixations on 1.54 million frames of 9,727 videos (frame resolution: $1280\times720$) in three months. We use the desk-mounted eye-tracker Tobii Pro Fusion (250 Hz) to collect the driver fixations with well-calibrated eye vision. \\
\faCaretRight\textbf{Annotation Details}: To avoid the eye fatigue of each subject, we divide 9,727 videos into 83 long videos with about 10 minutes each. In addition, we group similar accident types into one long video as much as possible to allow the subject to accumulate experience for capturing the dangerous object better. Each subject can only annotate one long video within one hour, and each long video is watched by all subjects, where at least one hour is maintained for resting the eyes for the next time watching. To obtain the gaze map, 
similar to DADA-2000 \cite{DBLP:conf/itsc/FangYQXWL19}, each frame's final gaze map is obtained using a Gaussian filter ($50\times50$ pixel kernel) to convolute all fixation points of subjects. To match the frame rate, we accumulate the fixation points in 250Hz frames of each subject to 30Hz frames.
\begin{figure*}[!t]
    \centering
\includegraphics[width=0.99\linewidth]{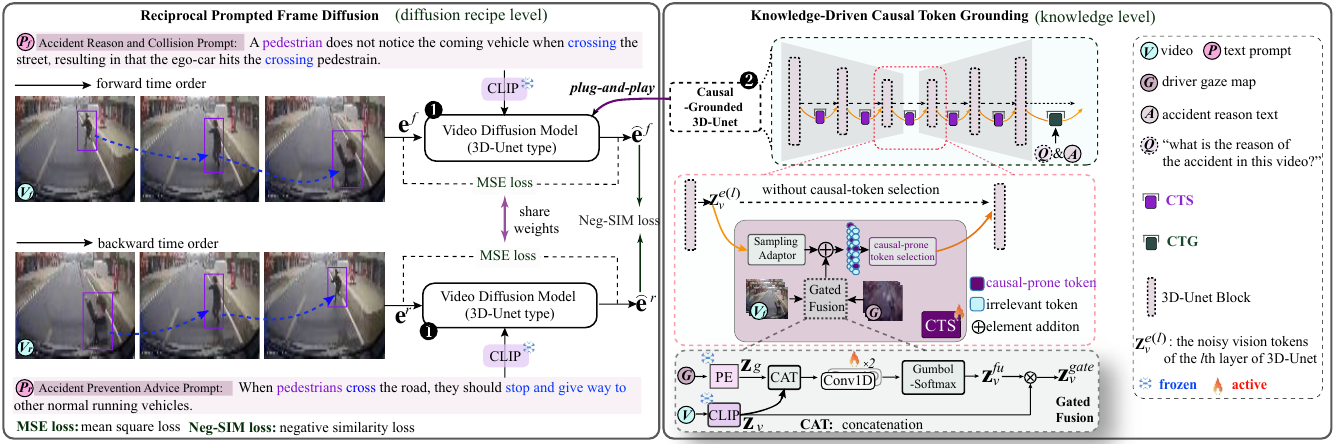}
    \caption{\small{The training schema of \textcolor{darkred}{\textbf{\modelname}}. It mainly includes three stages: The direct optimization of ${\bm{e}}^f$ with the forward time order diffusion (Stage-\textbf{0}, omitted in this figure), where the pre-trained Stable Diffusion~\cite{DBLP:conf/cvpr/RombachBLEO22} is used for model initialization. The reciprocal prompted frame diffusion (RPFD) in Stage-\ding{182} contrasts two diffusion pathways conditioned by reciprocal time order vision frames ($V_f$\&$V_r$) and semantically reciprocal text prompt ($P_f$\&$P_r$). We take the 3D-Unet backbone~\cite{DBLP:journals/corr/abs-2212-09381} in the noise (${\bm{e}}^f$\&${\bm{e}}^r$) representation learning at this stage. State-\ding{183} injects \emph{causal-prone token selection blocks} (\textcolor{darkred}{\textbf{CTS}}) and a \emph{causal token grounding} (\textcolor{darkblue1}{\textbf{CTG}}) block respectively into the inner layers and end layer of the 3D-Unet module to fulfill a causal-grounded video diffusion helped by accident reason ($Q$) answering ($A$) and gaze ($G$)-conditioned token selection modules. 
    }}
    \label{fig2}
    \vspace{-1em}
\end{figure*}\\
\faCaretRight\textbf{Utilization Ways}: We take all frames in \datasetname~for training use. Therefore, the video clips sampled in the diffusion process have pair-wise driver gaze maps and video frames. Additionally, \datasetname~can facilitate many egocentric traffic accident video understanding tasks, such as driver attention prediction~\cite{gan2022multisource,DBLP:conf/iccv/ChenNX23}; cognitive driving accident anticipation, like DRIVE~\cite{DBLP:conf/iccv/Bao0K21} and CogTAA~\cite{CognitiveTAA}; scanpath prediction in accident reason answering, like ~\cite{DBLP:conf/cvpr/ChenJZ21}, and other gaze map involved accident video understanding tasks.

\section{Methodology}

The essence of \modelname~is to fulfill a causal-grounded video content learning when providing certain accident-related text descriptions. To begin with, we analyze the video diffusion process in egocentric accident scenarios.

For an accident video clip $V$ coupling with an accident reason-collision description $P$, a noise representation ${\bm{e}}$$\sim $$\mathcal{N}(0, {\bm{I}})$ is leveraged for the video diffusion process. The \emph{forward process} $q({\bm{z}}_v^{e_k}|{\bm{z}}_v^{0})$ gradually adds ${\bm{e}}$ to ${\bm{z}}_v^{0}$ and generates the sequential-noisy latent vision variables ${\bm{z}}_v^{0}, {\bm{z}}_v^{e_2}, \dots, {\bm{z}}_v^{e_k}$ ($k$: diffusion step index), where 
\begin{equation}\small
{\bm{z}}_v^{e_k} = \sqrt{\hat{\beta}_{k}} {\bm{z}}_v^0 + \sqrt{1-\hat{\beta}_{k}} {\bm{e}}, 
\hat{\beta}_{k} = \prod_{i=1}^{k} \alpha_i, \alpha_k = 1 - \beta_k,
\end{equation}
$\beta_k$ is a parameterized schedule, and ${\bm{z}}_v^0$ is the pure video embedding of $V$. The \emph{reverse process} $q({\bm{z}}_v^{e_{k-1}}|{\bm{z}}_v^{e_k})$ recovers $\tilde{\bm{z}}_v^0$ by the following sampler step-by-step:
\begin{equation}
\tilde{\bm{z}}_v^{e_{k-1}} = \mu_{\hat{\bm{e}}}({\bm{z}}_v^{e_k}, k)+\sigma_k {\bm{e}},
\label{eq:mean}
\end{equation}
where
$\mu_{\hat{\bm{e}}}({\bm{z}}_v^{e_k}, k) = \frac{1}{\sqrt{\hat{\beta}_k}} \left( {\bm{z}}_v^{e_k} - \frac{\beta_k}{\sqrt{1 - \hat{\beta}_k}} \phi_{\hat{{\bm{e}}}}(k, {\bm{z}}_v^{e_k}, P) \right)$,
and $\sigma_k$ is the noise variance at the $k^{th}$ step ($k$ is omitted in following for simpilicity).
Let’s denote $\tilde{V}$ as the generated clip by a trained video diffusion model $\phi_{\hat{{\bm{e}}}}$ conditioned by $\{V, P, \hat{\bm{e}}\}$. The goal is to denoise the causal relations of latent variables ${\bm{z}}_v^{e}$ step-by-step from $V$ to $\tilde{V}$ and identify the causal-grounded (CG) latent vision representation ${\bm{z}}_v^{c}:{\bm{z}}_v^{e}|_\text{CG}$, responding well to the counterfactual change of text prompt $P$, optimized by the mean square loss (MSE):  
\begin{equation}
\text{min}_{\hat{\bm{e}}, {\bm{z}}_v^c}: \mathbb{E}_{{\bm{e}} \sim \mathcal{N}(0, {\bm{I}})} \lVert {\bm{e}} - \phi_{\hat{\bm{e}}} (k,{\bm{z}}_v^{e},P)\rVert_2^2.
\label{eq:mse}
\end{equation} 
Manifestly, learning the causal entity within accident videos (\emph{i.e.}, reasoning ${\bm{z}}_v^c$ associated with $P$) for video diffusion models is challenging because of the tiny and sudden change of causal entities. Hence, as shown in Fig.~\ref{fig2}, this work formulates \modelname~as two progressive levels:\\
\ding{182} \textbf{Diffusion recipe level} fulfills a contrastive forward and backward time order frame diffusion conditioned by semantically reciprocal text descriptions (\emph{abbrev., reciprocal prompted frame diffusion}), \emph{i.e.}, \textbf{forward}: accident reason and collision prompt; \textbf{backward}: accident prevention advice prompt. This recipe prefers to enhance the causal scene learning by global text-vision prompt intervention.\\
\ding{183}\textbf{Knowledge level} reforms the backbone (\emph{e.g.}, 3D-Unet) to be causal-grounded in causal token learning explicitly helped by accident reason answering head (ArA) and driver gaze map fusion with local token intervention.

\subsection {Reciprocal Prompted Frame Diffusion}
\label{recipe-level}
Reciprocal prompted frame diffusion (RPFD) (stage-\ding{182})is inspired by the greater attractiveness of the context of the critical scene or crash-prone objects than stable background scenes in egocentric accident videos~\cite{DBLP:conf/itsc/FangYQXWL19}. We leverage the hypothesis~\cite{DBLP:journals/corr/abs-2212-09381} that the accident prevention text prompt can coincide with the accident dissipation process reflected by backward time order accident frames. Differently, we argue that the backward-order diffusion can be treated as creating a counterfactual intervention to forward text and visual prompts, and the exogenous noise $\bm{e}$ helps the causal scene associate reciprocal frame and text prompts. 

With the reciprocal text and frame prompts, \textbf{different} text prompts should activate \textbf{different} visual content mainly associated with the causal entity in accident videos. 
Consequently, we contrast noise representation learning as
$\verb''$
\begin{equation}\small
\begin{aligned}
\mathcal{L_{\text{ST1}}} = \mathcal{L_{\text{MSE}}}({\bm{e}}_f,\hat{\bm{e}}_f) 
+ \mathcal{L_{\text{MSE}}}({\bf{e}}_r,\hat{\bm{e}}_r) 
+ \lambda \mathcal{L_{\text{NS}}}(\hat{\bm{e}}_f,\hat{\bm{e}}_r), \\
\mathcal{L_{\text{NS}}}(\hat{\bm{e}}_f,\hat{\bm{e}}_r) = \mathbb{E}\left(1-\frac{\hat{\bm{e}}_f\cdot \hat{\bm{e}}_r}{||\hat{\bm{e}}_f||||\hat{\bm{e}}_r||}\right), \verb'           '
\end{aligned}
\label{eq:4}
\end{equation}
where $\hat{\bf{e}}_f$ and $\hat{\bf{e}}_r$ denote the reconstructed noise representation in contrastive two diffusion pathways with shared weights. $\mathcal{L_{\text{NS}}}$ is the negative similarity loss (inverse cosine similarity) to contrastively enhance the different visual content learning in RPFD conditioned by forward prompt ($V_f$,$P_f$) and backward prompt ($V_r$,$P_r$). $\mathcal{L_{\text{MSE}}}$ takes the form of Eq.~\ref{eq:mse}, and $\lambda$=$0.2$ is a hyperparameter to balance the losses. 
\subsection{Knowledge-Driven Causal Token Grounding}
\label{model-level}
Reciprocal prompted frame diffusion (RPFD) aims to suppress the influence of scene background on the frame-level diffusion recipe. To find the causal-entity reflected regions, we further reform the diffusion backbone to be hierarchically causal-grounded in the Stage-\ding{183}. 
\begin{figure}[!t]
    \centering
\includegraphics[width=0.95\linewidth]{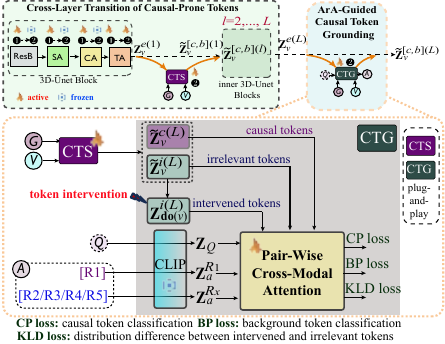}
    \caption{\small{The details of Cross-Layer Causal-Prone Token Transition and ArA-Guided Causal Token Grounding in the 3D-Unet backbone, where the \textcolor{darkblue1}{\textbf{CTG}} block are illustrated. When \textcolor{darkred}{\textbf{CTS}} and \textcolor{darkblue1}{\textbf{CTG}} are removed in inference, $\tilde{\bm{z}}_v^{[c,b](l)}$=$\bm{z}_v^{e(l)}$, marked by ``$\dashrightarrow$".}}
    \label{fig3}
    \vspace{-1.5em}
\end{figure}

As shown in Fig.~\ref{fig2} and Fig.~\ref{fig3}, the multi-layer 3D-Unet module in the diffusion model has multi-interleaved attention blocks, where we design \textcolor{darkred}{\textbf{CTS}} and \textcolor{darkblue1}{\textbf{CTG}} as flexible blocks that can be plug-and-play injected in the inner layers or the end layer of the 3D-Unet~\cite{DBLP:journals/corr/abs-2212-09381} structured by residual block (\textbf{ResB}), spatial (\textbf{SA}), cross-modal (\textbf{CA}), and temporal attention (\textbf{TA}) blocks. Concisely, we take \textcolor{darkred}{\textbf{CTS}} as a bridge to fulfill cross-layer \emph{causal-prone} token transition and involve an ArA head at \textcolor{darkblue1}{\textbf{CTG}} block to fulfill ArA-guided causal token grounding. Because of the sudden scene change in egocentric accident videos, we introduce the driver gaze to help the causal-prone token selection in \textcolor{darkred}{\textbf{CTS}}. Therefore, it is worth mentioning that \textcolor{darkred}{\textbf{CTS}} and \textcolor{darkblue1}{\textbf{CTG}} blocks can be seen as \emph{powerful guider} to refine causal tokens layer-be-layer, which are only used in the training phase and removed in the inference stage. \\
$\verb'  '$\textbf{Gating Allocation of Driver Gazes (Gated Fusion)}.
As illustrated in Fig.~\ref{fig3}, each \textcolor{darkred}{\textbf{CTS}}  receives the tokenized driver gaze map ${\bm{z}}_g$, visual tokens ${\bf{z}}_v$ of video frames $V$, and the noisy vision representation ${\bm{z}}_v^{e(l)}$ from previous temporal attention (TA) in each diffusion step, where $l$ indexes the inner layer of 3D-Unet module. Vision tokens ${\bm{z}}_v$ are encoded by a pre-trained frozen CLIP model~\cite{radford2021learning}. Because of the sparse distribution of driver fixations, driver gaze tokens ${\bm{z}}_g$ are obtained by only the position embedding (PE) layer in the same CLIP model to avoid the zero-value issue in CLIP's deep layers. Here, the PE layer is fulfilled by a 2D convolution (\emph{kernel size:} $14\times14$)). To avoid the influence of gaze bias in the collection, we provide a gated fusion of ${\bm{z}}_g$ and ${\bm{z}}_v$ by
\begin{equation}
\begin{aligned}
{\bm{z}}_{v}^{fu} &= \text{Gumbel-Softmax}(\text{Conv1Ds}(\text{cat}({\bm{z}}_v, {\bm{z}}_g))), \\
{\bm{z}}_{v}^{gate}  &= {\bm{z}}_v \otimes {\bm{z}}_{v}^{fu}, \verb''\otimes- \text{element multiplication},
\end{aligned}
\label{eq:5}
\end{equation}
where \text{Conv1Ds}(.) denotes two layers of 1D-convolution with \emph{relu} operation for reducing the token channel dimension after concatenation (\emph{i.e.}, cat(,.,)) of vision and gaze tokens. \text{Gumbel-Softmax}(.) ensures the values of ${\bm{z}}_{v}^{fu}$ at each token dimension summing to 1 for gated token selection~\cite{DBLP:conf/iclr/JangGP17}.\\
$\verb'  '$\textbf{Causal-Prone Token Selection (\textcolor{darkred}{\textbf{CTS}}) and Transition}.
In Stage-\ding{183}, the noisy vision representation ${\bf{z}}^e_v$ passes through multiple layers of the 3D-Unet module interleaved with SA, TA, and CA blocks. Therefore, reasoning the causal tokens ${\bm{z}}_v^{c}$ in ${\bm{z}}_v^{e}$ from the deep layer structure of 3D-Unet module is challenging. Hence, \textcolor{darkred}{\textbf{CTS}} blocks can be vividly understood as pulling out the causal tokens to be grounded layer by layer, where gaze maps are adopted to strengthen the pulling force by gated allocation.

\emph{Cross-Layer Token Sampling Adaptor}: As shown in Fig.~\ref{fig3}, we inject \textcolor{darkred}{\textbf{CTS}} block after each TA layer, helped by the temporal token transition ability. Assume the noisy vision representation outputted by the temporal attention (TA) layer is ${\bm{z}}_v^{e(l)}$ at the $l^{th}$ ($l=1,2,..., L$) inner layer of the 3D-Unet module, which is firstly processed through a \emph{sampling adaptor} for resizing it to adapt to the cross-layer Unet-scale change and match the CLIP output dimension~\cite{radford2021learning} adopted in gated fusion ${\bm{z}}_{v}^{gate}$, achieved by a bilinear interpolation (BintP) and a Conv2D ($1\times1$) operation as
\begin{equation}\small
\tilde{\bm{z}}_v^{e(l)}=\text{Conv2D}(\text{BintP}({\bm{z}}_v^{e(l)})),  \verb'' \hat{\bm{z}}_v^{\text{CP}(l)}=\tilde{\bm{z}}_v^{e(l)}\oplus{\bm{z}}_{v}^{gate},\\
\label{eq:6}
\end{equation}
where $\hat{\bm{z}}_v^{\text{CP}(l)}$ is the element addition $\oplus$ of $\tilde{\bm{z}}_v^{e(l)}$ and ${\bm{z}}_{v}^{gate}$. 

\emph{Causal-Prone Token Selection}: We compute the token importance score encoded from each video frame by $\emph{softmax}(\text{MLP}(\hat{\bm{z}}_{v}^{\text{CP}(l)}))$ with multilayers of perceptrons (MLP) and select tokens with top-$d$ scores, where we set $d$ empirically as a quarter of tokens within the single frame as the casual-prone tokens $\tilde{\bm{z}}_v^{c(l)}$ based on the long-tailed region distribution of causal objects and won't miss too many causal tokens (Here we omit the frame index for simplicity). The reminder ones are treated as the background tokens $\tilde{\bm{z}}_v^{b(l)}$. The combination $\tilde{\bm{z}}_v^{[c,b](l)}$ of $\tilde{\bm{z}}_v^{c(l)}$  and $\tilde{\bm{z}}_v^{b(l)}$ is fed into the ResB block in the next 3D-Unet block ($l$<$L$). If we encounter the
$L^{th}$ layer, ${\bm{z}}_v^{e(L)}$ is grounded by the following ArA head.\\
$\verb'  '$\textbf{ArA-Guided Causal Token Grounding (\textcolor{darkblue1}{\textbf{CTG}})}.
ArA-guided causal token grounding leverages the insight of the VideoQA task~\cite{DBLP:journals/pami/ChenZNZX23,liu2023cross,blubaum2024causal}, while differently we design \textcolor{darkblue1}{\textbf{CTG}} block in token-level grounding and prefer an ArA-guided video diffusion. The answers are multi-choice with the accurate one [$R_1$], and four disturbing ones [$R_2, R_3, R_4, R_5$] corresponding to the egocentric accident video clip [$V$].

As illustrated in Fig.~\ref{fig3}, if we receive ${\bm{z}}_v^{e(L)}$, it also involves the gate-conditioned \textcolor{darkred}{\textbf{CTS}} block and obtains $\tilde{{\bm{z}}}_v^{c(L)}$ and $\tilde{\bm{z}}_v^{b(L)}$. Differently, we take a token intervention on $\tilde{\bm{z}}_v^{b(L)}$ as $\tilde{\bm{z}}_{{\bm{do}}(v)}^{b(L)}$ by randomly masking a quarter of tokens to noise. With these token representations, the same cross-modal attention (CA) in 3D-Unet is adopted to classify the causal tokens and background tokens aligned by the text tokens of the accurate reason-question pair (${\bm{z}}_a^{R_1}$,${\bm{z}}_Q$) and disturbing accident reason-question token pairs $({\bm{z}}_a^{R_{x(x\neq1)}},{\bm{z}}_Q)$.

We employ cross-entropy loss (XE)~\cite{li2022invariant, li2022acmmm} for the causal and background token classification. Notably, the XE for background tokens follows a uniform distribution $\mathcal{U}(0,1)$ over all irrelevant answer candidates. The distribution difference between the intervened tokens $\tilde{\bm{z}}_{{\bm{do}}(v)}^{b(L)}$ and $\tilde{\bm{z}}_v^{b(L)}$ is measured by the Kullback-Leibler Divergence (KLD)~\cite{li2022invariant}. Thus, the loss function of ArA grounding is 
\begin{equation}
\mathcal{L}_{\text{ArA}}= \mathcal{L}_{\text{XE}}^{c} + \mathcal{L}_{\text{XE}}^{b} + \mathcal{L}^{{\bm{do}}(b)}_{\text{KLD}},
\label{eq:LST2_split}
\end{equation}
and 
{\small{$\mathcal{L}^{{\bm{do}}(b)}_{\text{KLD}} = \text{KLD}(\text{CA}({\bm{z}}^{R_x}_a|\tilde{{\bm{z}}}_v^{b(L)},{\bm{z}}_Q), \text{CA}({\bm{z}}^{R_x}_a|\tilde{{\bm{z}}}_{{\bm{do}}(v)}^{b(L)},{\bm{z}}_Q))$}}.   

\subsection{Training and Inference}

\faCaretRight\textbf{Training Recipe}: Stage-\textbf{0}, Stage-\ding{182}, and Stage-\ding{183} are progressively trained, and we use the same MSE loss in Eq. \ref{eq:mse} for noise representation learning. In Stage-\textbf{0}, only the forward time order video diffusion (\emph{i.e.}, only with ${\bm{e}}^f$) is conducted on the 3D-Unet backbone and trained 10,000 training steps. Stage-\ding{182} re-loads the pre-trained parameters in Stage-\textbf{0}, and fine-tunes the 3D-Unet module with contrastive noise learning (\emph{i.e.}, ${\bm{e}}^f$ and ${\bm{e}}^r$) by 10,000 steps, and then the trained parameters are reloaded in Stage-\ding{183} and are further trained 10,000 steps after adding \textcolor{darkred}{\textbf{CTS}} and \textcolor{darkblue1}{\textbf{CTG}} modules. In Stage-\ding{183}, the total loss involved the ArA head is: 
\begin{equation}
\mathcal{L}_{\text{ST2}}=\mathcal{L}_{\text{MSE}}({\bm{e}}^f,\hat{\bm{e}}^f)+\gamma \mathcal{L}_{\text{ArA}},
\label{eq:st1}
\end{equation}
where ${\bm{e}}^f\sim \mathcal{N}(0,{\bm{I}})$ is randomly re-initialized and $\gamma$=0.3 balances the weight of two terms.\\ 
\faCaretRight\textbf{Inference}: We provide the video-to-video (V2V) and text-to-video (T2V) synthesizing choices. Notably, \textcolor{darkred}{\textbf{CTS}} and \textcolor{darkblue1}{\textbf{CTG}} blocks are plug-and-play, which are removed in the inference stage, \emph{i.e.}, once two levels of fine-tuning \ding{182} and  \ding{183} are completed, the ArA head and driver gaze maps are removed. In the V2V mode, we can implement accident content editing and normal-to-accident video diffusion conditioned by the video clip input (\textcolor{darkred}{\textbf{$V$}}) and text prompt (\textcolor{blue}{\textbf{$P$}}). T2V mode starts from an initial random noise and generates the expected video frames conditioned by text prompt (\textcolor{blue}{\textbf{$P$}}). In the reverse diffusion process, we adopt the common DDIM scheduler~\cite{DBLP:conf/iclr/SongME21} to decode the video frames.   

\begin{figure*}[!t]
    \centering     \includegraphics[width=0.98\linewidth]{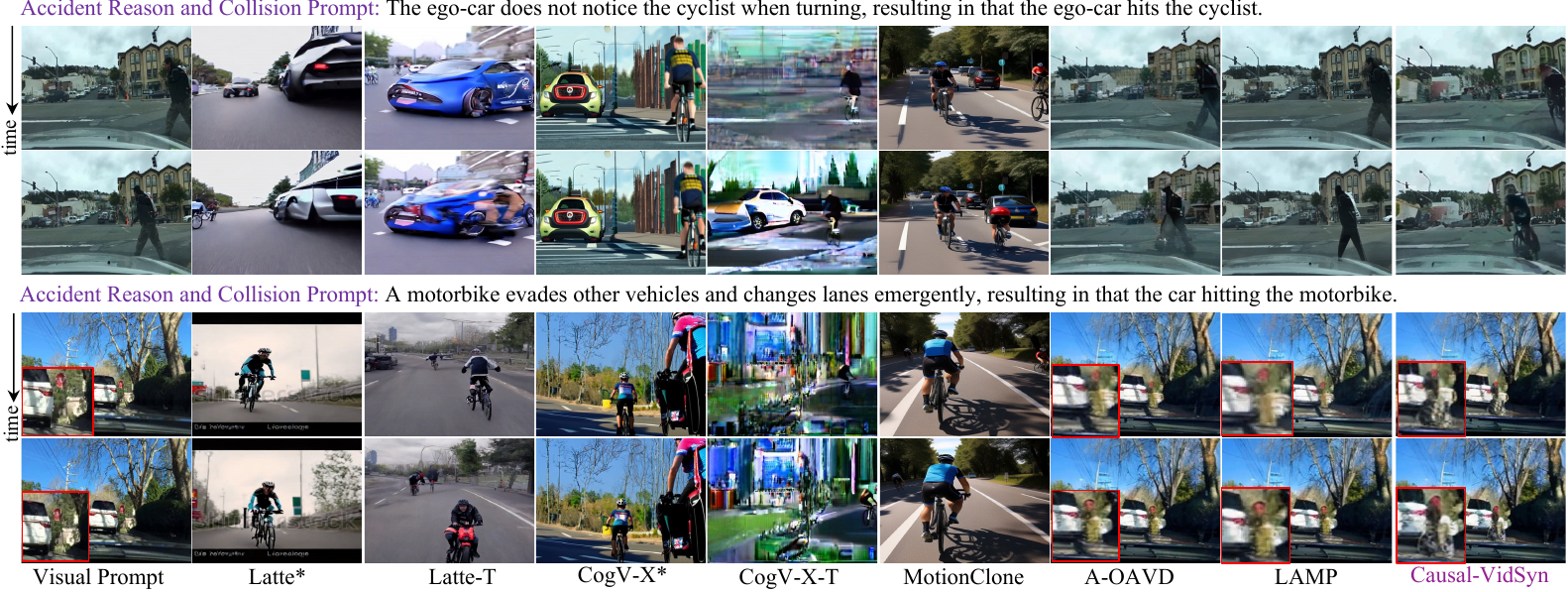}
        \vspace{-1em}
    \caption{\small{Sample visualizations of N2A task by Latte*~\cite{DBLP:journals/corr/abs-2401-03048}, Latte-T~\cite{DBLP:journals/corr/abs-2401-03048}, CogV-X*~\cite{yang2024cogvideox},  CogV-X-T~\cite{yang2024cogvideox}, MotionClone~\cite{DBLP:journals/corr/abs-2406-05338}, A-OAVD~\cite{DBLP:journals/corr/abs-2212-09381}, LAMP~\cite{wu2024lamp}, and our \modelname~(Best viewed in zoom mode).}}
    \label{fig4}
    \vspace{-1em}
 \end{figure*}
\section{Experiments}
\subsection{Tasks and Datasets}
We extensively evaluate the performance by three promising video generation tasks in testing: 1) normal-to-accident video diffusion (\textbf{N2A}), 2) accident video content editing (\textbf{AEdit}), and 3) text-to-accident video generation (\textbf{T2V}). Because the accident commonly appears suddenly with a very short time window~\cite{DBLP:journals/corr/abs-2212-09381}, we generate 16 frames to show the near-crash to crashing (NC-2-C) process concisely.\\
\faCaretRight\textbf{N2A} is a first-launched task in this field, which can be used to create labels for the detection of crash-prone objects and accident anticipation on accident-free AV testing platforms, \emph{e.g.,} nuScenes~\cite{DBLP:conf/cvpr/CaesarBLVLXKPBB20} and accelerate self-driving car testing. This work chooses a safety-critical dataset BDD-A~\cite{DBLP:conf/accv/XiaZKNZW18} to evaluate whether the critical scene context is aligned with the accident-related texts. We sample the end 16 frames (appearing critical objects) of 2,000 videos as visual prompts.\\
\faCaretRight\textbf{AEdit} task checks the causal-sensitivity by tiny text changes (\emph{e.g.}, ``\emph{pedestrian}"$\rightarrow$ ``\emph{cyclist}") reflecting the causal-entities. In this task, the background scenes are preferred to be unchanged. In this task, we adopt the DADA-2000 dataset~\cite{DBLP:journals/tits/FangYQXY22}, which also provides the driver fixations for checking whether the noticed objects aligned with text prompts are edited. We sample 3,000 clips in DADA-2000 from the NC-2-C frame windows in inference. \\
\faCaretRight\textbf{T2V} is well-known and we generate 500 clips conditioned by accident reason-collision text prompts in DADA-2000.\\
\faCaretRight\textbf{Training Dataset}. We sample 6,492 clips from the NC-2-C windows in 9,727 accident videos of \textbf{MM-AU}~\cite{DBLP:journals/corr/abs-2212-09381} with pair-wise accident reason-collision and prevention text descriptions, randomly sampled from the near-crash to crashing frame windows to maintain diversity.
\subsection{Implementation Details}
All the experiments of \modelname~and other Unet-based diffusion models are conducted using two NVIDIA RTX3090 GPUs with each 22GB RAM. The frame resolution is 224$\times$224. The learning rate of each stage is $1e-5$ using the Adam solver with $\beta_1$=$0.9$ and $\beta_2$=$0.999$.
We also involve Diffusion Transformer (DiT)-based methods for comparison. Because of the large frame resolution (CogVideoX-2B (CogV-X)~\cite{yang2024cogvideox}: 720$\times$ 480; Latte~\cite{DBLP:journals/corr/abs-2401-03048}: 512$\times$ 512) in diffusion, they are trained in one NVIDIA A800 GPU with 80GB RAM. The number of diffusion steps $k$ in each stage is set to $1,000$. The batch size for training \modelname~ is 2, and the layers $L$ of the 3D-Unet module is 12 with a symmetric structure.\\
\faCaretRight\textbf{Metrics}. Following the popular models~\cite{yang2024cogvideox,wang2025seedvrseedin}, we evaluate the frame quality by Fréchet video distance (\textbf{FVD}~\cite{DBLP:journals/corr/abs-1812-01717}), causal-sensitivity by clip score (\textbf{CLIP$_s$})~\cite{DBLP:journals/corr/abs-2212-09381,DBLP:journals/corr/abs-2405-14864}, frame content consistency by temporal consistency (\textbf{TempC}~\cite{DBLP:conf/iccv/QiCZLWSC23}).
\begin{table}[!t]\scriptsize
\centering
\setlength{\tabcolsep}{0.0mm}{
\begin{tabular}{l|ccc|cc|c}
 \toprule
\multirow{2}{*}{Methods} & \multicolumn{3}{c|}{\textbf{N2A} (BDD-A~\cite{DBLP:conf/accv/XiaZKNZW18} (2000))} & \multicolumn{2}{c|}{\textbf{T2V} (500)}  & \multirow{2}{*}{Backbone} \\
\cline{2-6}
& CLIP$_s$$\uparrow$ & FVD$\downarrow$& TempC$\uparrow$& CLIP$_s$$\uparrow$& TempC$\uparrow$\\
\hline\hline
T2V-zero~\cite{DBLP:conf/iccv/KhachatryanMTHW23}$_\text{CVPR2023}$    
& \underline{26.0} & 11754.8& 0.992& 24.8  &0.929&  Unet  \\ 
Free-bloom ~\cite{huang2024free}$_\text{NeurIPS2024}$   
& 25.1 & 8280.1& 0.990 &   22.1  & 0.841& Unet   \\
CoVideo~\cite{ControlVideo2024}$_\text{ICLR2024}$ 
& 24.2 & 9906.1& 0.930 & -& - & Unet   \\ 
Latte$^{*}$~\cite{DBLP:journals/corr/abs-2401-03048}$_\text{Arxiv2024}$ 
& 25.8 & 11066.2&0.993
& 22.8 & 0.977 & DiT \\  
CogV-X$^{*}$~\cite{yang2024cogvideox}$_\text{ICLR2025}$ 
&25.6 & 9075.6& 0.992 &  26.0  & \textbf{0.994}& DiT  \\
MotionClone~\cite{DBLP:journals/corr/abs-2406-05338}$_\text{ICLR2025}$ 
&24.9 &11554.9& \underline{0.994} &  -  & -& Unet  \\
\hline
TAV~\cite{DBLP:journals/corr/abs-2212-11565}$_\text{ICCV2023}$ 
& 24.1 & 9305.3 & 0.902
& 23.5 & 0.820& Unet \\ 
A-OAVD~\cite{DBLP:journals/corr/abs-2212-09381}$_\text{CVPR2024}$    
& 25.8 & 6378.9 & 0.992
& 26.5  &0.977& Unet     \\
LAMP~\cite{wu2024lamp}$_\text{CVPR2024}$  
& 25.4 & \underline{6208.2}&0.991&-& -& Unet  \\ 
Latte-T~\cite{DBLP:journals/corr/abs-2401-03048}$_\text{Arxiv2024}$ 
& 25.3 & 11196.6 &0.993
&25.6& 0.977 & DiT \\ 
CogV-X-T~\cite{yang2024cogvideox}$_\text{ICLR2025}$   
& 24.6 & 10094.0 &0.993
& 25.3 &\underline{0.981} & DiT  \\
AVD2~\cite{li2025avd2}$_\text{ICRA2025}$   
& - & - &-
& 27.3 &0.976 & DiT  \\
\hline
w/o [G+\textcolor{darkred}{\textbf{CTS}}\&\textcolor{darkblue1}{\textbf{CTG}}+RPFD] &25.8& 6378.9&0.992 &26.5 &0.976&Unet\\
 w/o [G+\textcolor{darkred}{\textbf{CTS}}\&\textcolor{darkblue1}{\textbf{CTG}}] &25.0& 7225.5  &\textbf{0.997}&26.4 &0.949&Unet\\
  w/o [G] &26.0&6249.3&0.992 &\underline{27.4}&0.945&Unet\\
\textcolor{darkred}
{\textbf{\modelname~[Full Train]} }   
& \textcolor{darkred}{\textbf{26.5}} & \textcolor{darkred}{\textbf{6192.3}}& \textcolor{darkred}{\underline{0.994}}
& \textcolor{darkred}{\textbf{27.5}}  & \textcolor{darkred}{0.944 }&
Unet \\
\hline
\end{tabular}}
\begin{tablenotes}
\item \scriptsize{The numbers in the brackets denote the sample scale in inference.}
\end{tablenotes}
\vspace{-1em}
\caption{\small{Performance on N2A and T2V tasks (\textbf{bold} font: the best).}}
\label{tab2}\vspace{-2em}
\end{table}
\faCaretRight\textbf{Competitors}. In comparison, we choose many SOTA video diffusion models because of their training efficiency and their ability to generate high-visual quality videos. They are divided as training-free methods, \emph{i.e.}, ControlVideo (\emph{Abbrev.}, CoVideo)~\cite{ControlVideo2024}, T2V-zero~\cite{DBLP:conf/iccv/KhachatryanMTHW23}, Free-bloom ~\cite{huang2024free}, Latte*~\cite{DBLP:journals/corr/abs-2401-03048}, CogVideoX*(\emph{Abbrev.}, CogV-X*)~\cite{yang2024cogvideox}, and MotionClone~\cite{DBLP:journals/corr/abs-2406-05338},
and training methods including Tune-A-Video (\emph{Abbrev.}, TAV)~\cite{DBLP:journals/corr/abs-2212-11565}, Abductive-OAVD (\emph{Abbrev.}, A-OAVD)~\cite{DBLP:journals/corr/abs-2212-09381}, LAMP~\cite{wu2024lamp} and the trained version of Latte-T and CogVideoX-T (\emph{abbrev.,} CogV-X-T). The training-free ones solely leverage their pre-trained models and directly infer the generation, and the training ones are fine-tuned 10,000 training steps with their official setting for involving more egocentric accident knowledge for fair comparison, via the forward time order frame diffusion by the same training set as our \modelname. Besides, we further involve one new work AVD2~\cite{li2025avd2} targeting  accident video synthesis (trained by MM-AU dataset~\cite{DBLP:journals/corr/abs-2212-09381}) in T2V evaluation.
\begin{table}[!t]\scriptsize
\centering
\setlength{\tabcolsep}{2mm}{
\begin{tabular}{l|cccc}
 \toprule
\multirow{2}{*}{Methods} & \multicolumn{4}{c}{\textbf{AEdit} (DADA~\cite{DBLP:journals/tits/FangYQXY22} (3000))} \\
\cline{2-5}
& CLIP$_s$$\uparrow$  & FVD$\downarrow$ & TempC$\uparrow$& \textbf{Afd}$\uparrow$ (\%)\\
\hline\hline 
TAV~\cite{DBLP:journals/corr/abs-2212-11565}$_\text{ICCV2023}$   
& 23.8 & 10076.2& 0.909 &-  \\ 
Latte-T~\cite{DBLP:journals/corr/abs-2401-03048}$_\text{Arxiv2024}$ 
&28.2 & 12377.3 & 0.945
&-\\
CogV-X-T~\cite{yang2024cogvideox}$_\text{ICLR2025}$   
& 25.3 & 11420.5 &0.945&- \\
A-OAVD~\cite{DBLP:journals/corr/abs-2212-09381} $_\text{CVPR2024}$      
& 26.9 & 5358.2 & 0.947 &49.4 \\
LAMP~\cite{wu2024lamp}$_\text{CVPR2024}$  
& 26.1 & 6191.6 & \textbf{0.971}&34.7  \\ 
\hline
 w/o [G+\textcolor{darkred}{\textbf{CTS}}\&\textcolor{darkblue1}{\textbf{CTG}}+RPFD] &26.9& 5358.8&0.947&47.1\\
 w/o [G+\textcolor{darkred}{\textbf{CTS}}\&\textcolor{darkblue1}{\textbf{CTG}}] &28.6& 6374.8  &0.942&50.3 \\
 w/o [G] &28.6&5368.2&0.939&50.4\\
\textcolor{darkred}
{\textbf{\modelname~[Full Train]}}    
& \textcolor{darkred}{\textbf{28.7}}& \textcolor{darkred}{\textbf{5352.9}}&\textcolor{darkred}{ 0.940}&\textcolor{darkred}{\textbf{55.4}}\\
\hline
\end{tabular}}
\vspace{-1em}
\caption{\small{The video generation performance of SOTA video diffusion models on the AEdit task ( \textbf{bold} font: the best).}}
\label{tab3}\vspace{-1em}
\end{table}
\begin{figure}[!t]
  \centering
\includegraphics[width=1.0\linewidth]{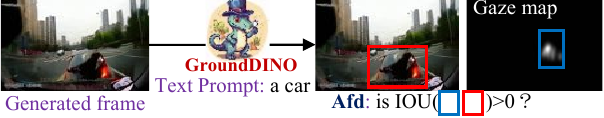}
   \vspace{-1em}
   \caption{\small{\textbf{Afd} is the ratio of $\text{IOU}(,)>0$ of all checks. IOU: the intersection over union of two bounding boxes.}}
   \label{fig5}
   \vspace{-1.5em}
\end{figure}
\subsection{Main Results on N2A and T2A Tasks}
\faCaretRight\textbf{N2A Task}. In this task, A-OAVD~\cite{DBLP:journals/corr/abs-2212-09381}, CogV-X*~\cite{yang2024cogvideox}, and our \modelname~are top three solutions showing better text-vision alignment than others, as shown in Tab.~\ref{tab2}. We also visualize the generated frames in Fig.~\ref{fig4}, and we can see that only our \modelname~can generate the critical cyclist while maintaining the frame style (the $1^{st}$ example in Fig.~\ref{fig4}). In addition, our \modelname~can generate an expected collision while maintaining the background reflected in the $2^{nd}$ sample of Fig.~\ref{fig4}. 
State-of-the-art methods, \emph{i.e.}, Latte*~\cite{DBLP:journals/corr/abs-2401-03048}, CogV-X*~\cite{yang2024cogvideox}, and MotionClone~\cite{DBLP:journals/corr/abs-2406-05338} generate clear frames with active response, while the frame background is not maintained with irrelevant object styles. After involving more accident knowledge into CogV-X-T (CogV-X*$\rightarrow$CogV-X-T), it generates artifacts mainly because more accident knowledge is required for fitting its large-scale DiT-based parameters. Notably, the objects in the generated frames of CogV-X*, Latte-T, and MotionClone are very large (named as \textbf{large object issue}) which causes large but unreasonable CLIP$_s$ values.\\
\faCaretRight\textbf{T2V Task}. Because there is no video frame reference, the T2V task here is to evaluate the ability for semantic alignment in text-to-video generation, and the quantitative results are focused here. From Tab.~\ref{tab2}, 
our \modelname~is the best for semantic alignment, and CogV-X* generates the best temporal consistency while the CLIP$_s$ value has a large gap to our model. Qualitative visualizations on the T2V task are shown in the supplementary file.
\subsection{Main Results on AEdit Task}
To measure the portions of cases where the expected
objects are identified and edited in the video frames (causal-entity reflected), we introduce a new metric, affordance (\textbf{Afd}), that stems from Add-it image editor~\cite{nguyen2024Addit}. We present Fig.~\ref{fig5} to display the computing way of Afd. Notably, we follow \cite{nguyen2024Addit} and adopt the GroundDINO~\cite{DBLP:conf/eccv/LiuZRLZYJLYSZZ24} to fulfill the text-guided object detection, while we utilize the gazed regions instead of the object bounding boxes to match a human-preferable causal-entity editing checking based on the positive perception of critical objects by human attention~\cite{DBLP:conf/itsc/FangYQXWL19,DBLP:conf/iclr/LiuHL023}.

\begin{figure}[!t]
    \centering    
    \includegraphics[width=0.98\linewidth]{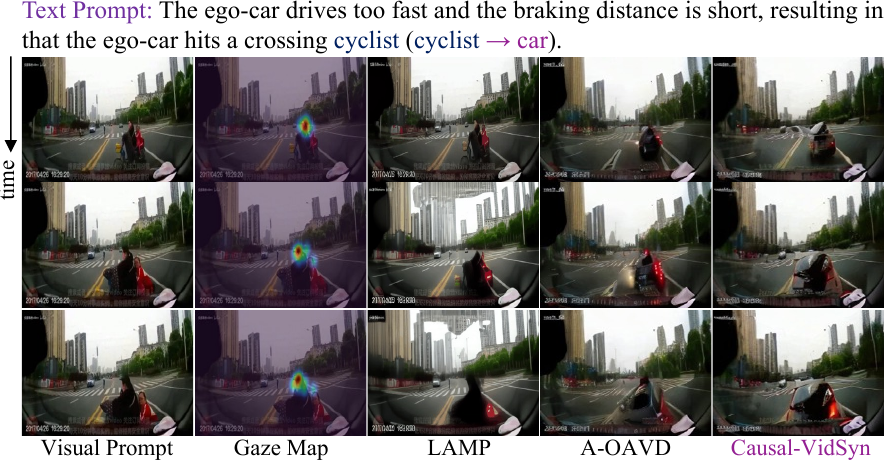}
    \vspace{-0.5em}
    \caption{\small{We visualize AEdit results of one crossing situation by LAMP~\cite{wu2024lamp}, A-OAVD~\cite{DBLP:journals/corr/abs-2212-09381}, and our \modelname.}}
    \label{fig6}
    \vspace{-0.5em}
\end{figure}

Based on the comparison in the N2A task, we can see that only LAMP, A-OAVD, and our \modelname~can maintain the frame style and the background scene context. Although Latte-T generates a high CLIP$_s$ (28.2) in the AEdit task, it is mainly caused by the \emph{large object issue}, as shown by the results in Tab.~\ref{tab3}. In this task, we visualize one example of DADA-2000 in Fig.~\ref{fig6}, where we change the ``cyclist'' to ``car'' in the text prompt and display the driver gaze maps to show the attentive objects. We can see that A-OAVD and our model show active response while the car shape is clearer in \modelname. LAMP~\cite{wu2024lamp} with a motion-consistency constraint can maintain the background style while it generates distorted and blurred content. From the \textbf{Afd} scores in Tab.~\ref{tab3}, our \modelname~can identify the fixed and crash-prone objects better and show stronger causal-sensitive object editing ability than LAMP and A-OAVD.\\
\begin{figure*}
    \begin{minipage}{\textwidth}
        \begin{minipage}[t]{0.6\textwidth}
            \begin{center}
              \includegraphics[width=\linewidth]{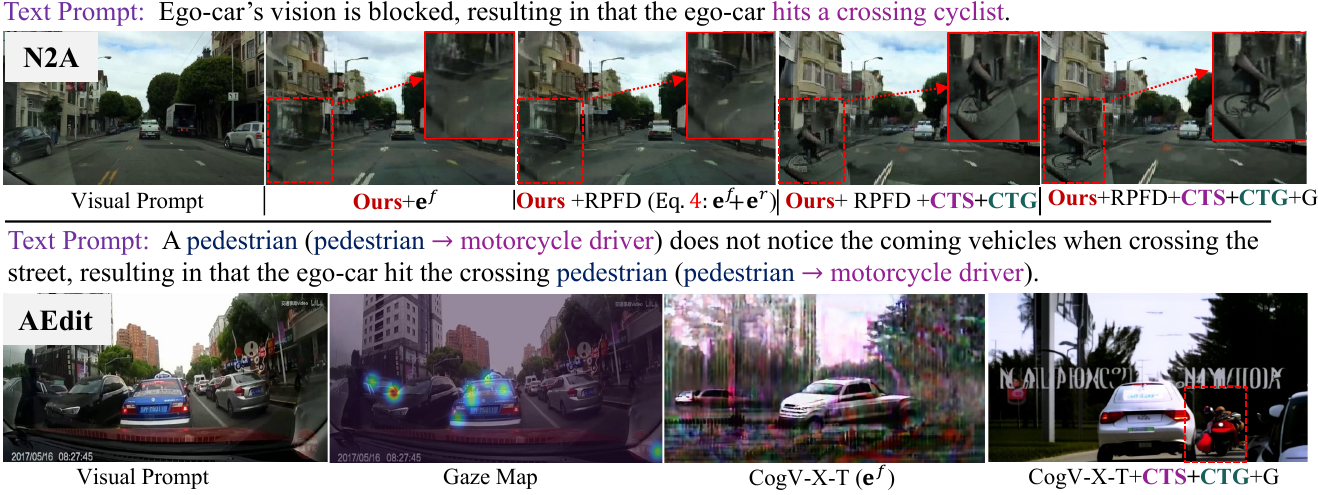}
            \end{center}
            \vspace{-18pt}           \caption{\small{Visualizations of one N2A example and an AEdit sample with the ablation checking of \modelname~and CogV-X-T~\cite{yang2024cogvideox}.}}
                \label{fig7}
        \end{minipage}
        \hspace{0.5ex}
        \begin{minipage}[t]{0.4\textwidth}\scriptsize
            \vspace{-1.53in}
            \centering
            \setlength\tabcolsep{8pt}
\setlength{\tabcolsep}{0.0mm}{
\begin{tabular}{l|cc|c|c}
\toprule
\multirow{2}{*}{Methods}  & \multicolumn{2}{c|}{\textbf{AEdit}} & \textbf{T2V} & \multirow{2}{*}{Tun.-Params}\\
\cline{2-4}
& CLIP$_s$$\uparrow$  & FVD$\downarrow$& CLIP$_s$$\uparrow$ \\
\hline\hline
TAV~\cite{DBLP:journals/corr/abs-2212-11565} (Unet)
& 23.8 & 10076.2
& 23.5 &  - \\ 
$\verb' '$+RPFD 
& 25.7 & 8701.4
& 23.7 &  0.24B (26\%) \\
$\verb' '$+RPFD+\textcolor{darkred}{\textbf{CTS}}(-G)
\&\textcolor{darkblue1}{\textbf{CTG}}
&27.5 & 8694.3
& 24.6 & 0.24B (26\%) \\
$\verb' '$+RPFD+\textcolor{darkred}{\textbf{CTS}}(+G)
\&\textcolor{darkblue1}{\textbf{CTG}} 
&\textcolor{darkred}{\textbf{27.9}} & \textcolor{darkred}{\textbf{8567.5}}
& \textcolor{darkred}{\textbf{25.4}} &  0.24B (26\%) \\
\hline
Latte-T~\cite{DBLP:journals/corr/abs-2401-03048}  (DiT)
& 28.2 & 12377.3 
&25.6 & -\\ 
$\verb' '$+\textcolor{darkred}{\textbf{CTS}}(+G)
\&\textcolor{darkblue1}{\textbf{CTG}}
& \textcolor{darkred}{\textbf{28.5}} & \textcolor{darkred}{\textbf{11316.9}}
&\textcolor{darkred}{\textbf{26.3}}& 1.06B(100\%)
 \\
 \hline
CogV-X-T~\cite{yang2024cogvideox}  (DiT) 
& 25.3 & 11420.5 
& 25.3 &  - \\
$\verb' '$+\textcolor{darkred}{\textbf{CTS}}(+G)
\&\textcolor{darkblue1}{\textbf{CTG}}   
& \textcolor{darkred}{\textbf{28.2}} & \textcolor{darkred}{\textbf{10892.9}}
&\textcolor{darkred}{\textbf{29.2}}  & 0.05B (2.9\%)\\
\hline
\end{tabular}}
 \begin{tablenotes} 
\item \footnotesize{\emph{Tun.-Params: fine-tuned parameters (rate\%).}}
\end{tablenotes}
\vspace{-6pt}
\captionsetup{font=small}        \makeatletter\def\@captype{table}\makeatother\captionsetup{font=small}
            \caption{\small{The diagnostic evaluation of TAV~\cite{DBLP:journals/corr/abs-2212-11565}, Latte-T~\cite{DBLP:journals/corr/abs-2401-03048}, and CogV-X-T~\cite{yang2024cogvideox} on \textbf{AEdit} and \textbf{T2V} tasks, with the fine-tuning by different causal-aware modules.}}
            \label{tab4}             
        \end{minipage}
    \end{minipage}
    \vspace*{-12pt}
\end{figure*}
\vspace{-1.5em}
\subsection{Diagnostic Experiments}
\faCaretRight\textbf{Roles of RPFD (Eq.~\ref{eq:4}), \textcolor{darkred}{\textbf{CTS}}, \textcolor{darkblue1}{\textbf{CTG}}, and Gaze Maps}. We evaluate different causal-aware modules in Stage-\ding{182} and \ding{183}. We gradually dismantle the gaze maps (G), \textcolor{darkred}{\textbf{CTS}}\&\textcolor{darkblue1}{\textbf{CTG}}, and RPFD in the fine-tuning phase and re-train the model with the same training dataset. Tab.~\ref{tab2} and Tab.~\ref{tab3} show the ablation results for N2A-T2V tasks, and the AEdit task, respectively.
From these results, all components are positive and RPFD seems with a significant role in the AEdit task claimed by the increase of CLIP$_s$ value. Removing \textcolor{darkred}{\textbf{CTS}}\&\textcolor{darkblue1}{\textbf{CTG}} displays a significant degradation. The N2A sample (creating a ``crossing cyclist") in Fig.~\ref{fig7} further verifies the positive role of \textcolor{darkred}{\textbf{CTS}}\&\textcolor{darkblue1}{\textbf{CTG}}. Gaze maps can localize the critical objects, but the accident causes can better explain their behaviors.\\
\faCaretRight\textbf{Portability of \textcolor{darkred}{\textbf{CTS}}\&\textcolor{darkblue1}{\textbf{CTG}}}. We also graft the causal-aware modules to fine-tune three other SOTA methods, \emph{i.e.,} Unet-based TAV, DiT-based Latte and CogV-X. TAV follows the same fine-tuning stages of \modelname. For DiT-based ones, because they do not have ResB block and the Unet-scale change, the \textcolor{darkred}{\textbf{CTS}} block is taken as an extra bridge between previous TA and next SA block and the \textcolor{darkblue1}{\textbf{CTG}} block is also attached at the end of DiT structure (see more details in the supplement). Then, Latte-T and CogV-T are further fine-tuned with 20,000 steps. We gradually add each module and check the performance gains. Tab.~\ref{tab4} demonstrates the effectiveness of the \textcolor{darkred}{\textbf{CTS}}\&\textcolor{darkblue1}{\textbf{CTG}} clearly, especially for the AEdit task by CogV-X-T (CLIP$_s$ +2.9). Notably, although CogV-T generates artifacts as analyzed in Fig.~\ref{fig4}, \textcolor{darkred}{\textbf{CTS}}\&\textcolor{darkblue1}{\textbf{CTG}} can fine-tune it to recover the content structure clearly, as shown by the AEdit example in Fig.~\ref{fig7}.\\
\begin{table}[!t]\scriptsize
\centering
\vspace{-0.5em}
\resizebox{\columnwidth}{!} %
{
\begin{tabular}{l|cc|ccc}
\toprule
\multirow{2}{*}{Methods} & \multicolumn{2}{c|}{\textbf{N2A} (BDD-A(2000))} & \multicolumn{3}{c}{\textbf{AEdit} (DADA (3000))} \\
\cline{2-6}
 &  CLIP$_s$$\uparrow$ & FVD$\downarrow$  & CLIP$_s$$\uparrow$ & FVD$\downarrow$  & \textbf{Afd}$\uparrow$\\
\hline\hline
A-OAVD~\cite{DBLP:journals/corr/abs-2212-09381} &25.8&6378.9 &26.9&5358.2&49.4 \\
\hline
\textcolor{darkred}
{\textbf{\modelname~[Full Train]}}  &\textcolor{darkred}{\textbf{26.5}}&\textcolor{darkred}{\textbf{6192.3}} &\textcolor{darkred}{\textbf{28.7}}&\textcolor{darkred}{\textbf{5352.9}} &\textcolor{darkred}{\textbf{55.4}}\\
Downscale Layers w/o \textcolor{darkred}{\textbf{CTS}} &26.2&6449.6 &28.3&5476.4&49.9 \\
Upscale Layers w/o \textcolor{darkred}{\textbf{CTS}} &26.3& 6408.9  &28.3 &5449.4 &54.9\\
with only \textcolor{darkblue1}{\textbf{CTG}} &26.0& 6512.9  &26.9 &5507.8&50.2\\
\hline
\end{tabular}} 
\vspace{-1em}
\caption{\small{The ablation studies by our \modelname~,\emph{w.r.t.}, different injection layers of \textcolor{darkred}{\textbf{CTS}} on the 3D-Unet module.}}
\label{tab5}
\vspace{-1.5em}
\end{table}
\faCaretRight\textbf{Roles of Different Injection Layers of \textcolor{darkred}{\textbf{CTS}}}.
As aforementioned, \textcolor{darkred}{\textbf{CTS}} and \textcolor{darkblue1}{\textbf{CTG}} can be understood as pulling out the causal tokens to be grounded layer by layer. Therefore, we offer more analysis for which part of the layers in the 3D-Unet module is dominant for selecting the causal tokens. We remove the \textcolor{darkred}{\textbf{CTS}} blocks from the downscale layers and upscale layers in the training phase, respectively. They are named as ``Downscale Layers w/o \textcolor{darkred}{\textbf{CTS}}'' and ``Upscale Layers w/o \textcolor{darkred}{\textbf{CTS}}''. In addition, we also maintain the final \textcolor{darkblue1}{\textbf{CTG}} only and remove all \textcolor{darkred}{\textbf{CTS}} blocks in the inner 3D-Unet layers. 
From Tab.~\ref{tab5}, we can observe that removing \textcolor{darkred}{\textbf{CTS}} in downscale layers has more degree of performance degradation, which indicates that the \textcolor{darkred}{\textbf{CTS}} is more important in the downscale layers than upscale layers. This observation verifies: \emph{deeper is better for inserting the \textcolor{darkred}{\textbf{CTS}} key}.\\
\begin{figure}[!t]
  \centering
\includegraphics[width=\linewidth]{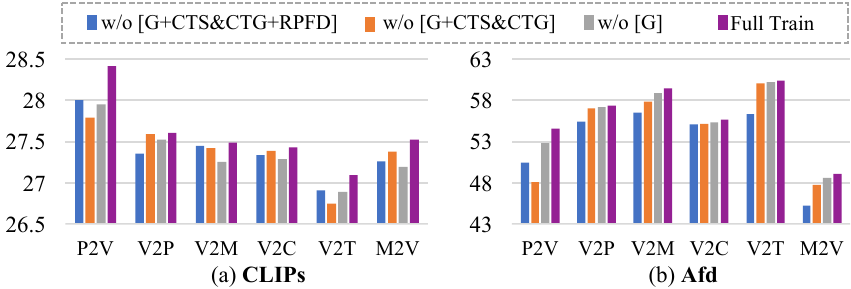}
   \vspace{-1em}
   \caption{\small{The CLIP$_s$ and Afd values of \modelname, \emph{w.r.t.}, fine-grained causal-entity editing from ``pedestrian'' to ``vehicle'' (P2V), ``vehicle'' to ``pedestrian'' (V2P),  ``vehicle'' to ``motorcycle/motorbike'' (V2M),  ``vehicle" to ``cyclist'' (V2C),  ``vehicle to ``truck'' (V2T),  and ``motorcycle/motorbike'' to ``vehicle'' (M2V).}}
   \label{fig8}
   \vspace{-1.5em}
\end{figure}
\faCaretRight\textbf{Fine-Grained Causal-Entity Editing}.
We present Fig.~\ref{fig8} for the analysis of the fine-grained causal-entity editing with the same ablation versions of \modelname~in Tab.~\ref{tab2} and Tab.~\ref{tab3}. From the statistics, it is clear that the driver gaze helps find causal-entity reflected by the sudden decreases of CLIP$_s$ scores when removing driver gaze assistance. In addition, for the ``vehicle'' to ``truck'' (V2T) group, the Afd values are larger than other groups, which is mainly caused by the large scale of the truck and can easily obtain a large IOU score (Fig.~\ref{fig5}). Additionally, the CLIP$_s$ values show inconsistent gain when removing different causal-aware modules on different groups. We think that this is caused by the text-vision semantic alignment contributed by the unmatched behaviors (not changed) of newly converted object types (different objects with distinct motion patterns). However, compared with the [Full Train] version, the causal-aware modules are positive for \modelname.
\section{Conclusions}
This work presents a new egocentric traffic accident video diffusion model \modelname, which enables the causal grounding in video diffusion via leveraging the cause descriptions and driver fixations to identify the accident participants and behaviors facilitated by accident reason answering and gaze-conditioned token selection modules (\textcolor{darkred}{\textbf{CTS}}\&\textcolor{darkblue1}{\textbf{CTG}}). The extensive experiments and analysis show the designed \textcolor{darkred}{\textbf{CTS}}\&\textcolor{darkblue1}{\textbf{CTG}} are plug-and-play with clear effectiveness and portability and make \modelname~with better frame quality and causal-sensitivity for N2A, AEdit, and T2V tasks. In the future, we will investigate video diffusion for other accident understanding tasks, such as accident anticipation.



{
    \small
    \bibliographystyle{ieeenat_fullname}
    \bibliography{AAVD}
}
\clearpage
\setcounter{equation}{0}
\setcounter{figure}{0}
\setcounter{table}{0}
\setcounter{page}{1}
\setcounter{section}{0}

\twocolumn[{
\renewcommand\twocolumn[1][]{#1}
\maketitle
\begin{center}
    \textbf{\Large Supplementary Material of Causal-Entity Reflected Egocentric Traffic Accident Video Synthesis}
    \vspace{20pt}
     \centering
    \captionsetup{type=figure}
\includegraphics[width=\linewidth]{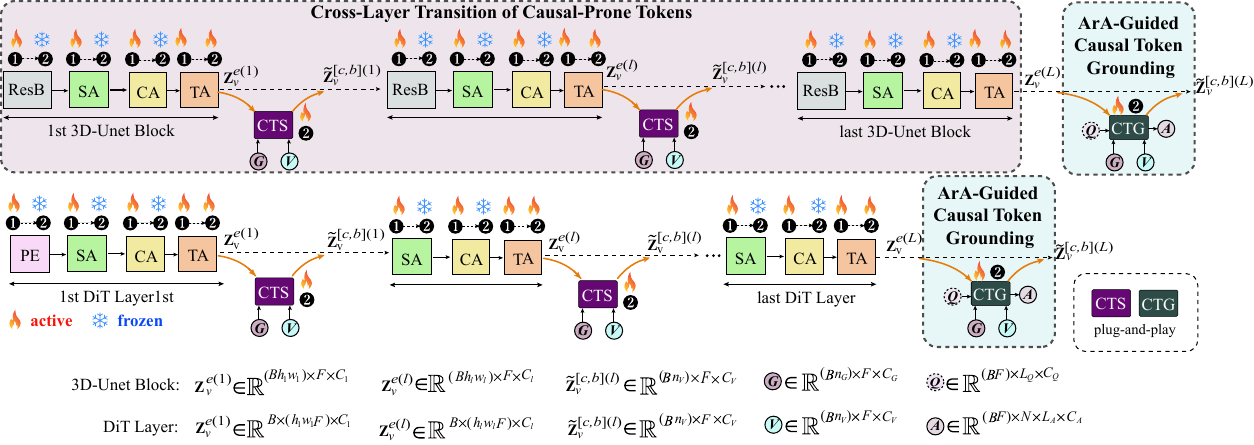}
\vspace{-1.8em}
   \caption{\small{\textbf{The detailed injection workflow of \textcolor{darkred}{\textbf{CTS}} and \textcolor{darkblue1}{\textbf{CTG}}} in \modelname, interleaved with multiple SA, CA, and TA blocks in 3D-Unet or DiT-based video diffusion models. The dimension indexes are defined as: $B$ denotes the batch size, $F=16$ is the video frame length, the maximum question prompt length $L_Q$ is set to 10, and the maximum answer prompt length $L_A$ is set to 32. In each layer of 3D-Unet block or DiT, $C_l$, $h_l$, and $w_l$ represent the channels, height, and width of the input noisy vision tokens ${\bm{z}}_v^{e(l)}$. $C_V$, $C_G$, $C_Q$, and $C_A$ are the token channels of video $V$, gaze maps (G), the question ($Q$) prompt and optional answers ($A$), respectively. $n_V$ and $n_G$ represent the number of vision tokens and gaze map tokens. Here, we set $C_V=C_G =C_Q=C_A=1024$ and $n_V=n_G=256$ in this work.}}
   \label{fig9}
\end{center}}]
\maketitle

\section{More Details of Implementation}

To be clear for re-reproduction, we provide the injection workflow of \textcolor{darkred}{\textbf{CTS}} and \textcolor{darkblue1}{\textbf{CTG}} in Fig.~\ref{fig9}, where different attention modules, \ie, \textbf{SA}, \textbf{CA}, and \textbf{TA}, are fed into low-rank adaptation (LoRA) trainer \footnote{\url{https://github.com/cloneofsimo/lora}} for fast fine-tuning. In Stage-\ding{183}, we mainly freeze other modules and only fine-tune TA, \textcolor{darkred}{\textbf{CTS}} and \textcolor{darkblue1}{\textbf{CTG}}, for Tune-A-Video (TAV)~\cite{DBLP:journals/corr/abs-2212-11565}, \underline{Latte-T+\textcolor{darkred}{\textbf{CTS}}+\textcolor{darkblue1}{\textbf{CTG}}+G}\footnote{The official Latte does not provide T2V training code, we take the implementation of \url{https://github.com/lyogavin/train_your_own_sora} for Latte's T2V training.}~\cite{DBLP:journals/corr/abs-2401-03048}, and our \modelname. For \underline{CogV-X-T+\textcolor{darkred}{\textbf{CTS}}+\textcolor{darkblue1}{\textbf{CTG}}+G}, we follow the official training strategy of CogV-X~\cite{yang2024cogvideox}, and update the parameters of query (\textbf{q}), key (\textbf{k}), value (\textbf{v}) of CA and TA blocks. 

\section{More Details of \textcolor{darkred}{\textbf{CTS}} and \textcolor{darkblue1}{\textbf{CTG}}}

\subsection{The Architecture of Sampling Adapter}
As denoted in Fig.~\ref{fig9}, ${\bm{z}}_v^{e(l)} \in \mathbb{R}^{(Bh_lw_l)\times F\times C_l}$. To match the dimension of ${\bm{z}}_v^{gate}$ in Eq.~\ref{eq:4} stated in the main paper, a bilinear interpolation (BintP) is applied to adjust the token dimension as $\mathbb{R}^{(Bh_pw_p)\times F \times C_l}$, where the size of $h_p\times w_p$ equals the number of vision tokens $n_V$. Then, we apply a 2D convolution (kernel size: $1\times1$) to transform $C_l$ to $C_V$. We reshape the token shape and output $\tilde{\bm{z}}_v^{e(l)}\in\mathbb{R}^{(Bn_V)\times F \times C_V}$, which is then fed into the \textcolor{darkred}{\textbf{CTS}} block.  

From Fig.~\ref{fig9}, it is worth noting that there is a change between the input and output of \textcolor{darkred}{\textbf{CTS}} block in the spatial dimension of tokens though the \emph{token sampling adapter} (Eq.~\ref{eq:6}). This aims to enhance the versatility of the \textcolor{darkred}{\textbf{CTS}} block, \emph{i.e.}, ensuring the noise representation $\bm{z}_v^e$ of different diffusion models (\emph{e.g.}, Unet- or DiT-based) to adapt to the video representation $\bm{z}_v$ (Eq.~\ref{eq:5}) outputted by CLIP model~\cite{radford2021learning} in  \textcolor{darkred}{\textbf{CTS}} block after token sampling adapter (Eq.~\ref{eq:6}). Actually, this is universal in the cross-attention (CA) module of 3D-Unet backbones to fulfill the text-vision alignment with the extra input of vision or text tokens. We also have attempted to keep the resolution of $\bm{z}_v$ (Eq.~\ref{eq:6}) consistent with $\bm{z}^e_v$ (Eq.~\ref{eq:mse}), while multi-layer \textcolor{darkred}{\textbf{CTS}} block will increase memory usage and computation cost exhaustively without versatility. 

Therefore, we try our best to minimize the influence of \textcolor{darkred}{\textbf{CTS}} on the spatial relationships of the tokens in the cross-layer transition within the 3D-Unet backbone as much as possible. As shown in Fig.~\ref{fig9}, the $1^{st}$ \textcolor{darkred}{\textbf{CTS}} block is injected at the end of the $1^{st}$ layer of 3D-Unet (DiT: the same way) backbone. In the inference phase, \textcolor{darkred}{\textbf{CTS}} and \textcolor{darkblue1}{\textbf{CTG}} are removed, where the output of the $1^{st}$ layer of 3D-Unet (or DiT) backbone is directly fed into the next layer of backbones. Additionally, only the temporal attention (TA) is fine-tuned in Stage-\ding{183}, which has not influence on the position indices of spatial attention (SA) and crossing attention (CA) modules. Therefore, in the inference phase, the spatial relationships of tokens are not disrupted.

   \begin{figure*}[htpb]
  \centering
\includegraphics[width=\linewidth]{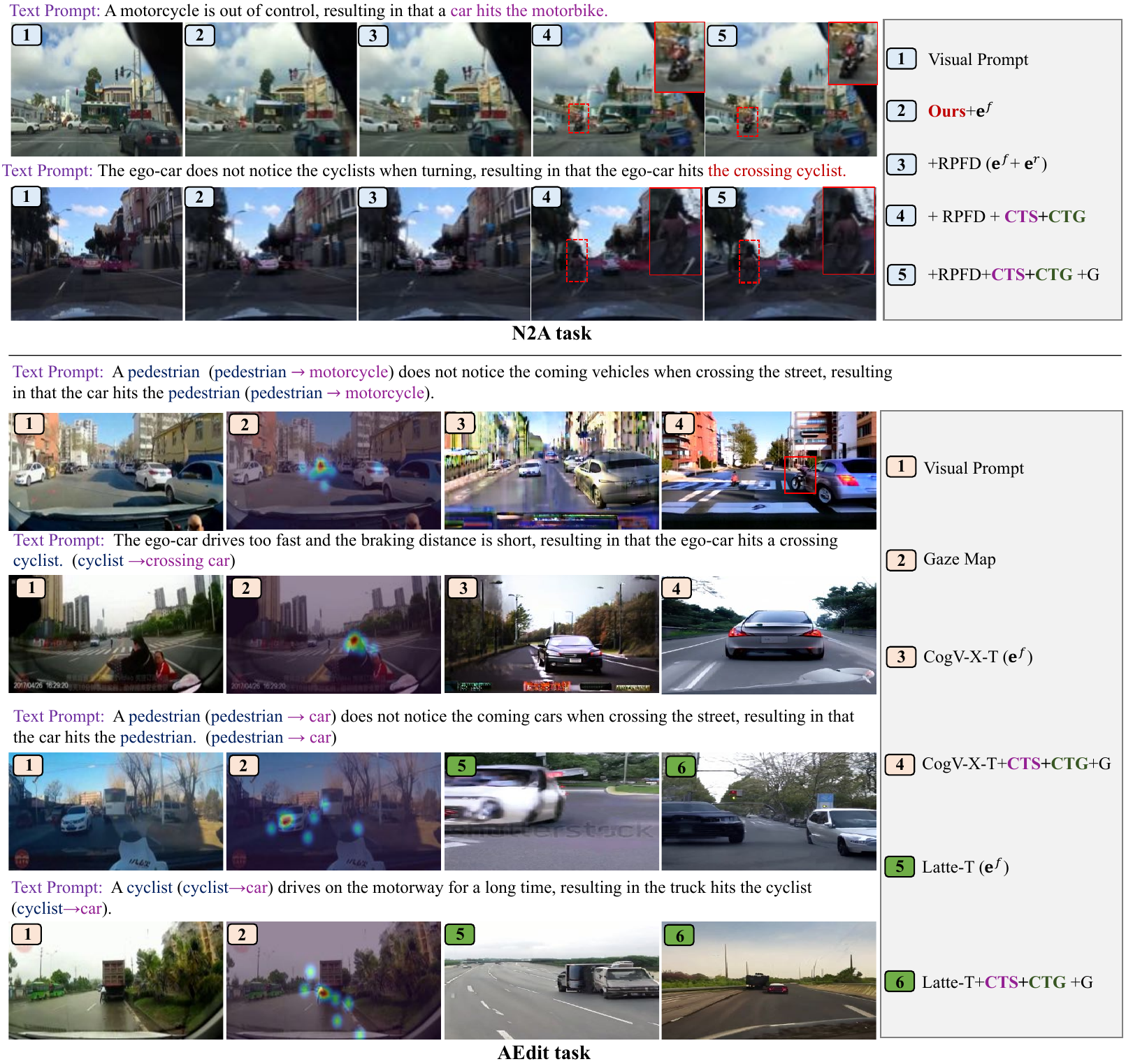}
   \caption{\small{We visualize two N2A examples and four AEdit samples with the ablation response checking of our \modelname~, CogV-X-T~\cite{yang2024cogvideox}, and Latte-T~\cite{DBLP:journals/corr/abs-2401-03048} with different causal-aware fine-tuning stages.}}
   \label{fig14}
   \vspace{-0.5em}
\end{figure*}
\subsection{Extending \textcolor{darkred}{\textbf{CTS}} and \textcolor{darkblue1}{\textbf{CTG}} to DiT-type VDMs}
As for the injection of \textcolor{darkred}{\textbf{CTS}} and \textcolor{darkblue1}{\textbf{CTG}} in DiT-based video diffusion models (VDMs), as shown in Fig.~\ref{fig9}, we denote the noisy vision tokens in the inner DiT layers as ${\bm{z}}_v^{e(l)}\in \mathbb{R}^{(B \times (h_l w_l F) \times C_l}$. Because there is no downscale or upscale operation in DiT layers, different from 3D-Unet, the sampling adapter just needs to reshape ${\bm{z}}_v^{e(l)}$ to match the dimension of ${\bm{z}}_v^{gate}$ without the BintP and Conv2D operations in 3D-Unet. Then, the gated fusion, \textcolor{darkred}{\textbf{CTS}} and \textcolor{darkblue1}{\textbf{CTG}} take the same structures.

To offer more evidence for the \textcolor{darkred}{\textbf{CTS}} and \textcolor{darkblue1}{\textbf{CTG}} in this work, we visualize more samples in Fig.~\ref{fig14} for checking their roles, where only the last frame in each generated clip is presented concisely. It can be observed that \textcolor{darkred}{\textbf{CTS}} and \textcolor{darkblue1}{\textbf{CTG}} modules can help CogV-X-T and Latte-T to recover the frame content (\emph{e.g.}, tree, building, and sky regions) in original video frames and present active response to the changed text phrase (cyclist/pedestrian$\rightarrow$ car). As for our \modelname, \textcolor{darkred}{\textbf{CTS}} and \textcolor{darkblue1}{\textbf{CTG}} perform well for active response corresponding to the given text prompt. In summary, the \textcolor{darkred}{\textbf{CTS}} and \textcolor{darkblue1}{\textbf{CTG}} modules can fine-tune the video diffusion models to identify the critical object effectively and rule out the influence of background scenes.

\begin{figure*}[htpb]
  \centering
\includegraphics[width=\linewidth]{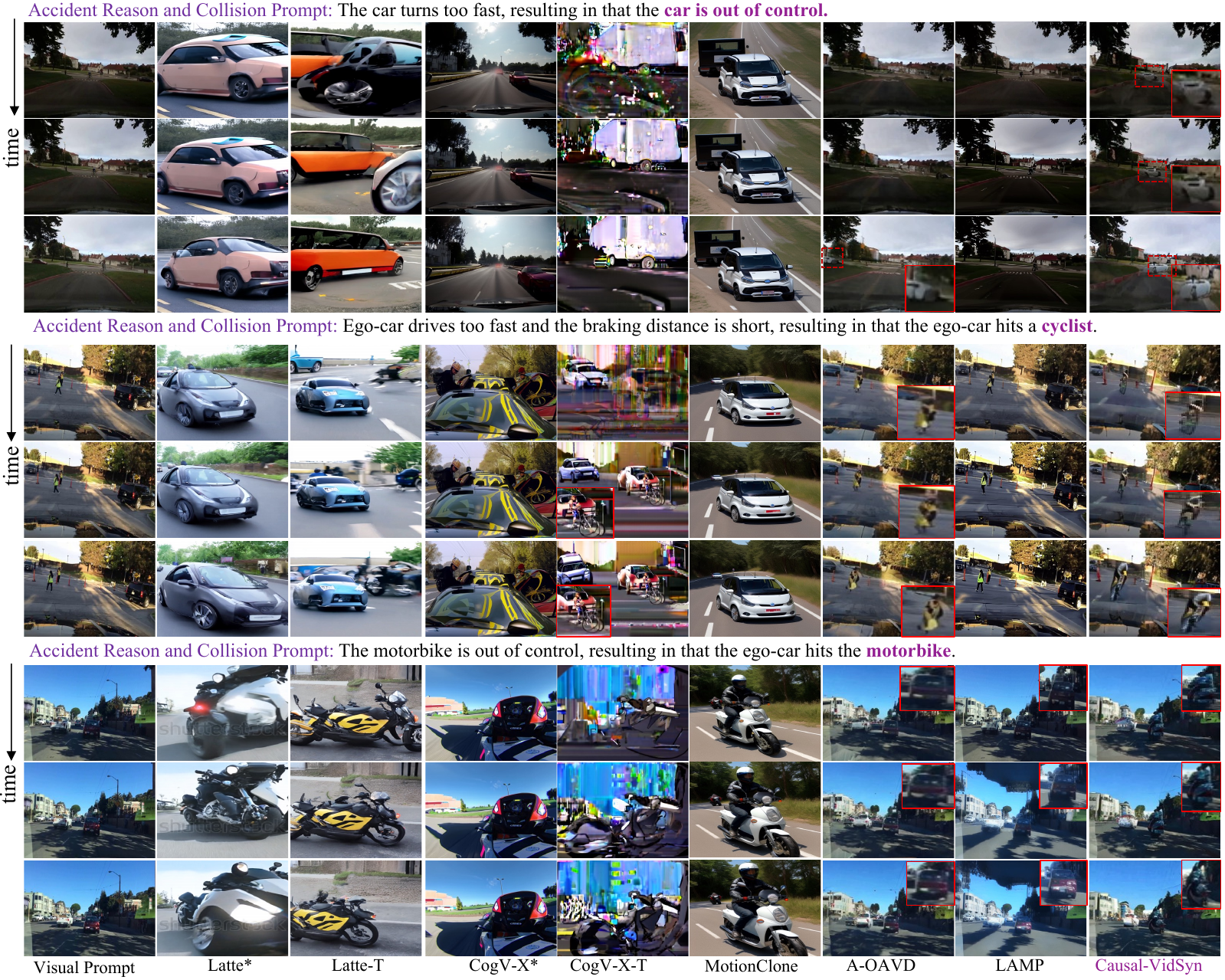}
  \caption{\small{Sample visualizations of N2A task by Latte*~\cite{DBLP:journals/corr/abs-2401-03048}, Latte-T~\cite{DBLP:journals/corr/abs-2401-03048}, CogV-X*~\cite{yang2024cogvideox},  CogV-X-T~\cite{yang2024cogvideox}, MotionClone~\cite{DBLP:journals/corr/abs-2406-05338}, A-OAVD~\cite{DBLP:journals/corr/abs-2212-09381}, LAMP~\cite{wu2024lamp}, and our \modelname~(Best viewed in zoom mode).}}
   \label{fig10}
\end{figure*}

   \begin{figure*}[htpb]
  \centering
\includegraphics[width=\linewidth]{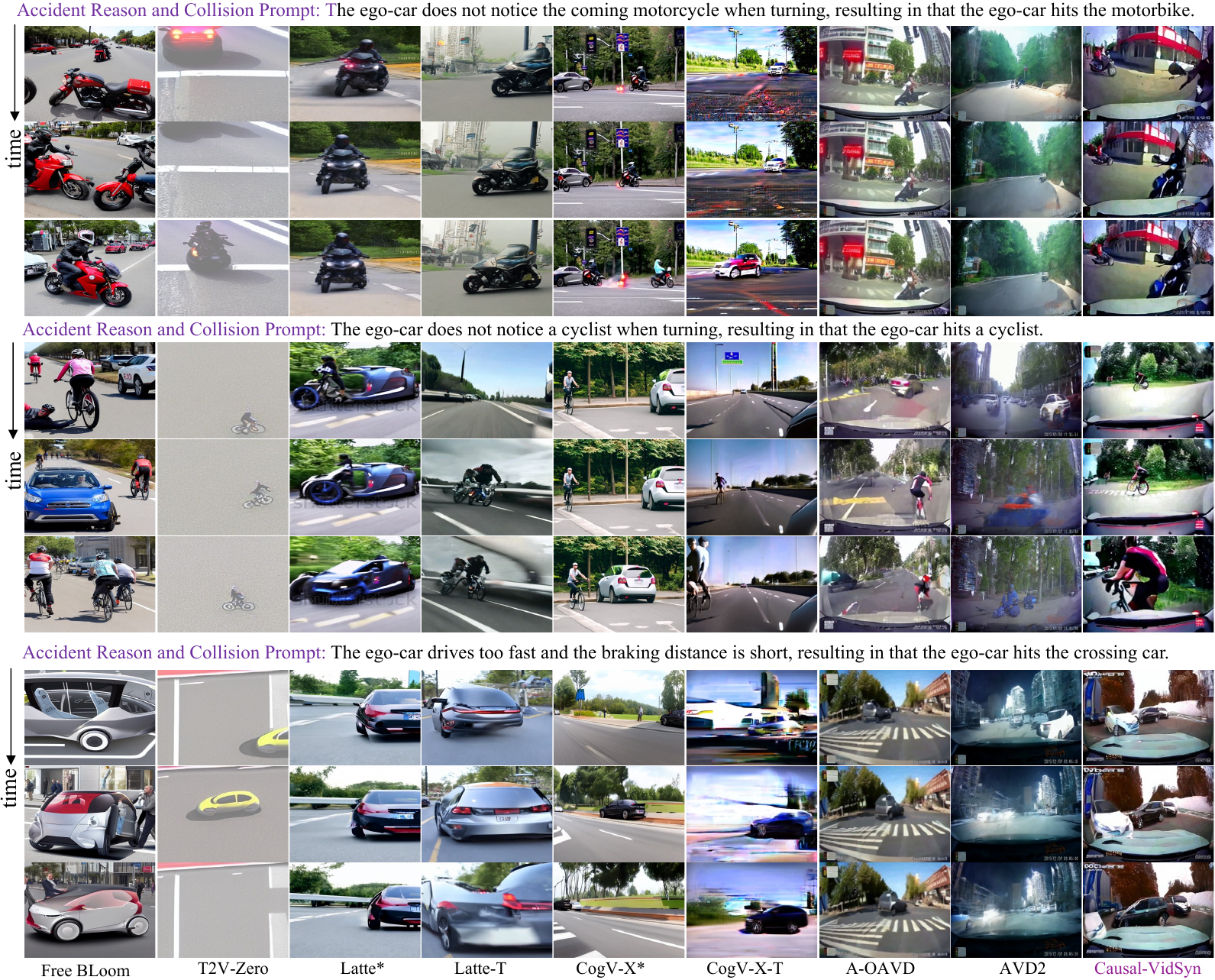}
  \caption{\small{Sample visualizations of T2V task by Free-bloom ~\cite{huang2024free}, T2V-Zero~\cite{DBLP:conf/iccv/KhachatryanMTHW23}, Latte*~\cite{DBLP:journals/corr/abs-2401-03048}, Latte-T~\cite{DBLP:journals/corr/abs-2401-03048}, CogV-X*~\cite{yang2024cogvideox},  CogV-X-T~\cite{yang2024cogvideox}, A-OAVD~\cite{DBLP:journals/corr/abs-2212-09381}, AVD2~\cite{li2025avd2}, and our \modelname~ (Best viewed in zoom mode).}}
   \label{fig12}
\end{figure*}

    \begin{figure*}[htpb]
  \centering
\includegraphics[width=\linewidth]{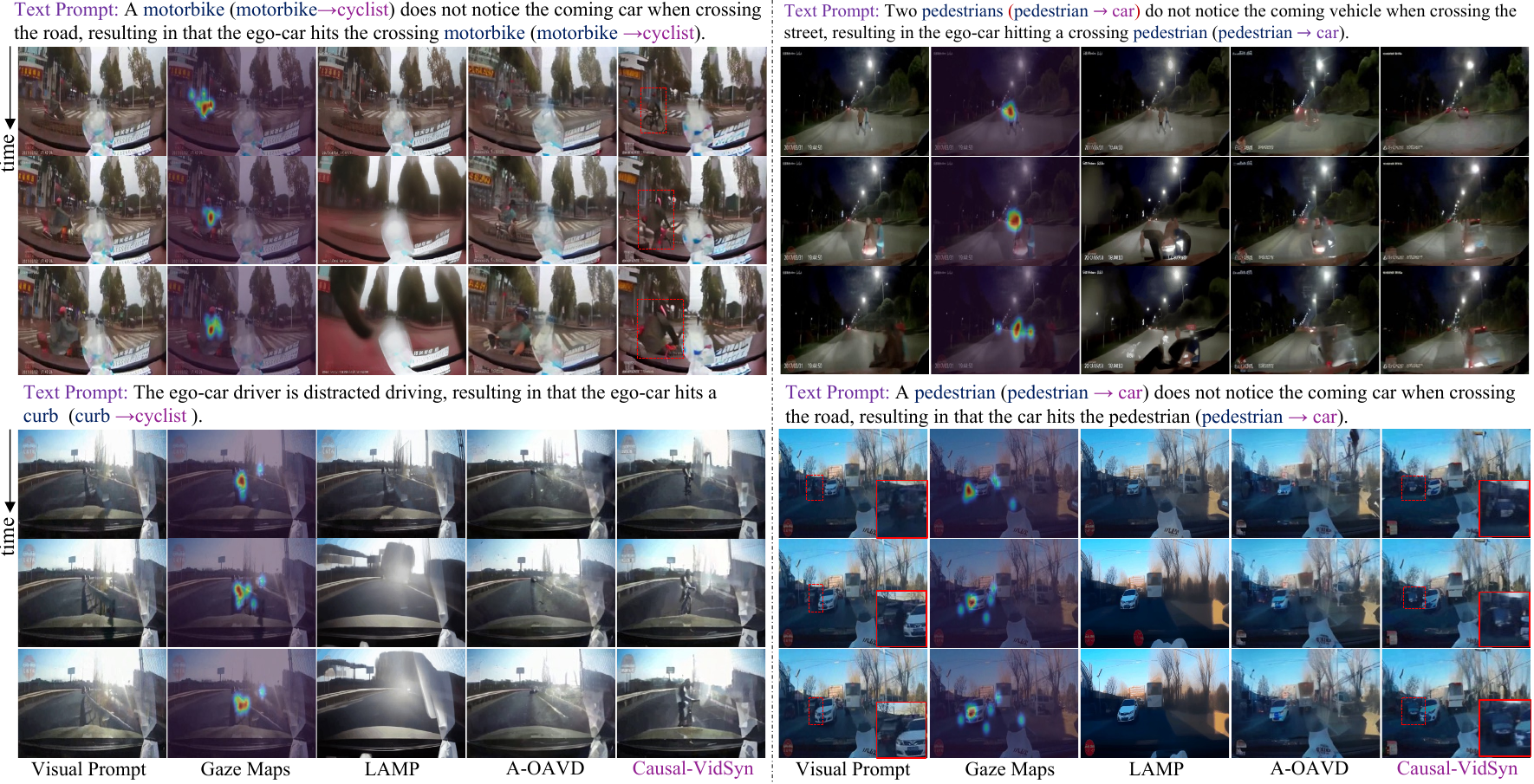}
   \caption{\small{We visualize more AEdit results of four different situations (rainy, dark, strong light, and shadow) by LAMP~\cite{wu2024lamp}, A-OAVD~\cite{DBLP:journals/corr/abs-2212-09381}, and our \modelname.}}
   \label{fig13}
\end{figure*}

\section{More Evaluations of \modelname}
For a solid evaluation, we offer more evaluations mainly from the visualizations of N2A, T2V, and AEdit tasks.
\subsection{More Visualizations on N2A and T2V Tasks}
\textbf{N2A Evaluation}: We present more ego-car involved visualizations of the N2A task in Fig.~\ref{fig10}. It can be observed that the ``large object issue'' is manifest in MotionClone~\cite{DBLP:journals/corr/abs-2406-05338}, and DiT-based methods, \emph{i.e.}, Latte*~\cite{DBLP:journals/corr/abs-2401-03048} and CogV-X*~\cite{yang2024cogvideox}, despite of the clear video frames. Additionally, the frame style changes dramatically for Latte* and CogVideoX*. After training by the egocentric accident videos (\emph{i.e.}, CogV-X*$\rightarrow$CogV-X-T), the content structure (\emph{e.g.}, the trees and sky) in the generated frames becomes closer to the original visual prompt. MotionClone is not consistently active for the given text prompt, such as the ``cyclist'' of the $1^{st}$ sample in Fig.~\ref{fig10}, and generates an irrelevant car. A-OAVD~\cite{DBLP:journals/corr/abs-2212-09381} and LAMP~\cite{wu2024lamp} show failures on the presented samples where the expected content change in most examples does not appear, such as the cyclist and motorbike in the $2^{nd}$ and $3^{rd}$ samples in Fig.~\ref{fig10}. Our \modelname~can change the critical objects with the best response, \emph{w.r.t.}, the given text prompt, while maintaining the frame background well.

\textbf{T2V Evaluation}: 
We present more T2V visualizations on ego-car involved accidents in Fig.~\ref{fig12}, where we compare Free-bloom ~\cite{huang2024free}, T2V-Zero~\cite{DBLP:conf/iccv/KhachatryanMTHW23}, Latte*~\cite{DBLP:journals/corr/abs-2401-03048}, Latte-T~\cite{DBLP:journals/corr/abs-2401-03048}, CogV-X*~\cite{yang2024cogvideox},  CogV-X-T~\cite{yang2024cogvideox}, A-OAVD~\cite{DBLP:journals/corr/abs-2212-09381}, AVD2~\cite{li2025avd2}, and our \modelname. From these visualization results, we can see that Free-bloom~\cite{huang2024free} likely generates multiple and similar objects while the collision situations do not occur. The results of T2V-Zero ~\cite{DBLP:conf/iccv/KhachatryanMTHW23} show more surveillance views. AVD2~\cite{li2025avd2}, fine-tuned Sora\footnote{\url{https://openai.com/sora/.}} by MM-AU dataset~\cite{DBLP:journals/corr/abs-2212-09381} can generate a similar frame style, while the expected collision is not well exhibited. In the T2V task, we can observe that the CogV-X* and Latte* and the other training-free methods cannot reflect the collisions well. As for A-OAVD, the expected near-crash scenarios appear while the collisions are only generated by our \modelname. From these results, it concludes that egocentric traffic accident knowledge is limited in these text-to-video generation methods.

\subsection{More Visualizations on the AEdit Task}
The AEdit task is directly to show the ability for causal-entity reflected accident video synthesis by counterfactual text prompt change. In Fig.~\ref{fig13}, we present four samples with the comparison of LAMP~\cite{wu2024lamp}, A-OAVD~\cite{DBLP:journals/corr/abs-2212-09381}, and our \modelname. LAMP~\cite{wu2024lamp} prefers collision-free and commonly focuses on the consistency of background scenes. A-OAVD~\cite{DBLP:journals/corr/abs-2212-09381} presents active responses while the target shape is not complete and the locations of the created targets have a larger distance to the ego-car compared with our \modelname. When facing strong or dark light conditions, A-OAVD cannot edit the expected video content well, \emph{w.r.t.}, the counterfactual text phrase. As for our \modelname, it shows a better causal-sensitive object editing, \emph{w.r.t.}, counterfactual text modification, than other models. 

  \begin{figure*}[htpb]
  \centering
\includegraphics[width=\linewidth]{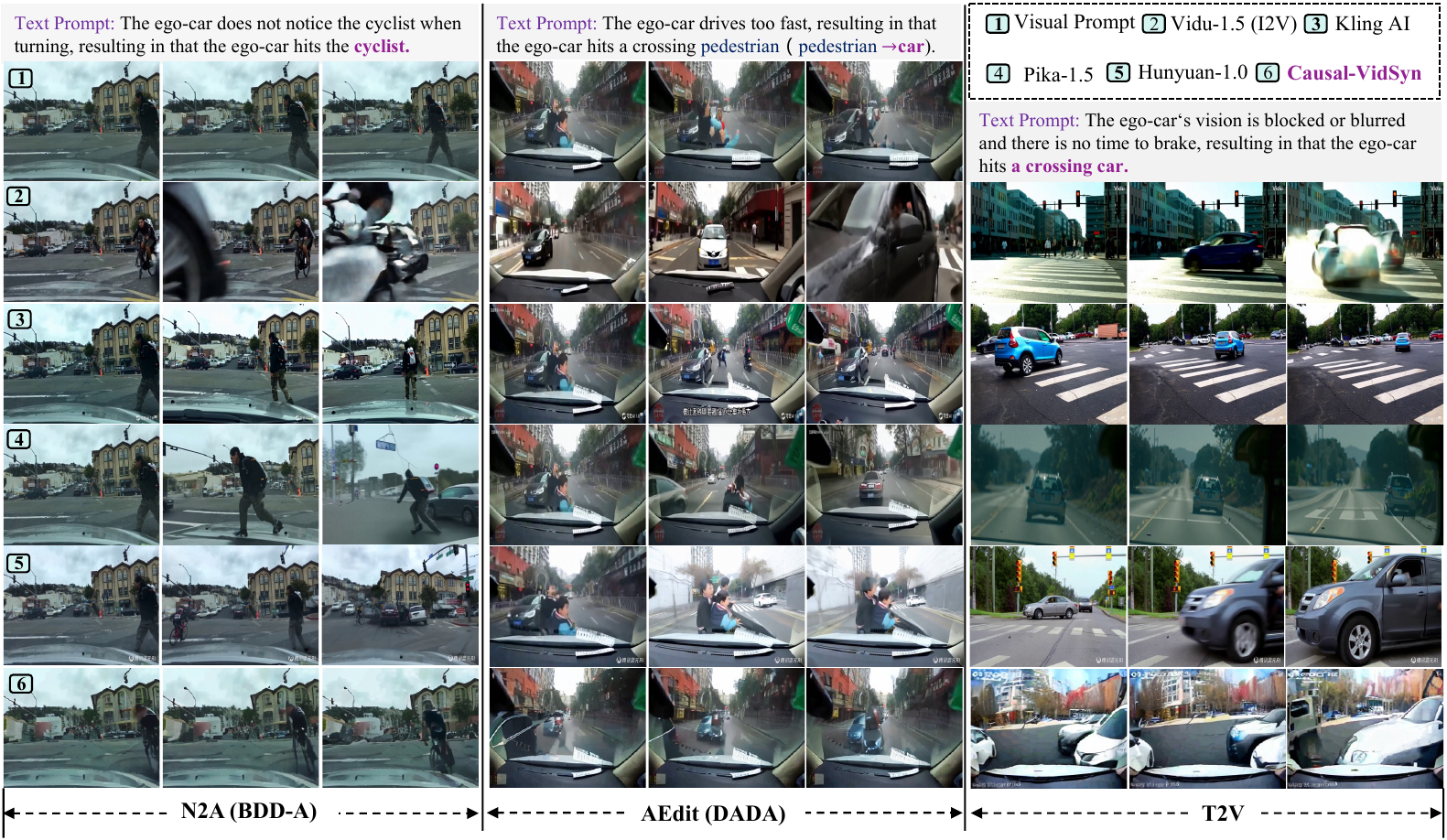}
   \caption{\small{Comparisons between \modelname~and four commercial model: Pika-1.5, Vidu-1.5 (I2V), Kling-AI, and HunyuanVideo-I2V.}}
   \label{fig15}
\end{figure*}

\section{Comparison with Commercial Models}
To verify the SOTA performance of our \modelname, we also take four popular and famous commercial models, including Pika-1.5\footnote{Release time: Oct 2, 2024, \url{pika.art/}}, Vidu-1.5\footnote{Release time: Nov 13, 2024, \url{vidu.studio/zh}}, Kling AI\footnote{Release time: Jul 25, 2024, \url{klingai.kuaishou.com/}}, and newly released HunyuanVideo-I2V\footnote{Release time: Mar 06, 2025, \url{https://github.com/Tencent/HunyuanVideo}}, for egocentric traffic accident video generation. Fig.~\ref{fig15} displays one sample for N2A, AEdit, and T2V tasks, respectively. From the results, it is interesting that only Vidu-1.5 conditioned by images (\emph{abbrev.,} Vidu-1.5 (I2V)) can generate egocentric accidents well for the examples in all tasks. The text prompt-driven vision change is not well exhibited for Pika-1.5, Kling-AI, and HunyuanVideo-I2V for the N2A and AEdit tasks. For example, the pedestrian in the first example in Fig.~\ref{fig15} is not changed to an expected cyclist.  Vidu-1.5 (I2V) can reflect the behavior or visual content change, while it does not display an active response of ``pedestrian''$\rightarrow$``car''. HunyuanVideo-I2V generates a good generation for the T2V task, while in N2A and AEdit tasks, it does not show the expected video content change. For Kling-AI, it contrarily shows an accident dissipation process, where the near-crash objects go far away from the ego-car given the text prompt. As for our \modelname, the expected egocentric accident situation is outputted. Certainly, these comparisons cannot represent all situations, but they can exhibit the causal sensitivity of our model for accident video content editing and egocentric accident video generation.

\begin{figure}[t]
  \centering
\includegraphics[width=\linewidth]{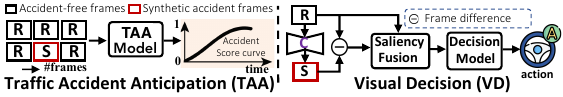}
   \vspace{-1.5em}
   \caption{\small{The TAA and decision Tasks. \textbf{R} is the real data, and \textbf{S} is the synthesis data driven by \textbf{C} (our~\textcolor{darkred}{\modelname}~model).}}
   \label{fig19}
\end{figure}
\section{Downstream Task Explorations}

We explore two downstream tasks using our synthetic data: traffic accident anticipation (TAA) and visual decision (VD). Fig.~\ref{fig19} shows the pipelines of them. In the TAA task, we take Cog-TAA~\cite{CognitiveTAA} as the baseline, and take the accident-free BDD-A~\cite{DBLP:conf/accv/XiaZKNZW18} datasets to synthesize accident frames by LAMP, OAVD, and our model (\textbf{C}), respectively, obtaining the same-scale accident and accident-free sample pairs to Cog-TAA for training.

\begin{figure}
    \begin{minipage}{\linewidth}
          \begin{minipage}[t]{0.65\textwidth}\scriptsize
            \centering
            \setlength\tabcolsep{8pt}
\setlength{\tabcolsep}{0.3mm}{
\begin{tabular}{l|ccc}
\toprule
{Baseline} & \multicolumn{3}{c}{DADA-2000 [\textcolor{green}{16}]}\\
\cline{2-4}
 & AP$\uparrow$ & AUC$\uparrow$ & TTA$_{0.5}$$\uparrow$ \\
\hline\hline
Cog-TAA [\textcolor{green}{36}] w/o \emph{T\&Att} &0.701 &\textbf{0.774}&3.742\\
\hline
Cog-TAA (LAMP) & 0.698 & 0.504 &3.210  \\
Cog-TAA (OAVD) & 0.735 & 0.654 &3.746 \\
Cog-TAA (Ours) & \textcolor{darkred}{\textbf{0.758}} & \textcolor{darkred}{\underline{0.689}} &\textcolor{darkred}{\textbf{3.762}}\\
\hline
\end{tabular}}
\vspace{-1.4em}
\captionsetup{font=small}        \makeatletter\def\@captype{table}\makeatother\captionsetup{font=small}
\caption{\small{The TAA task results.}}
\label{tab1}             
\end{minipage}
\hspace{0.5ex}
    \begin{minipage}[htpb]{0.3\linewidth}
            \begin{center}
              \includegraphics[width=\linewidth]{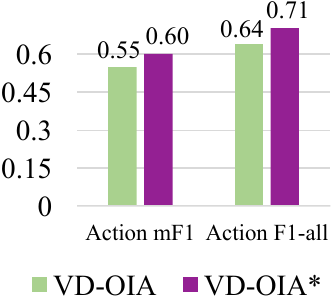}
            \end{center}
            \vspace{-18pt}           \caption{\small{The VD task results.}}
                \label{fig20}
                  \vspace{-2em}   
        \end{minipage} 
    \end{minipage}
\end{figure}

The same DADA-2000 test set~\cite{DBLP:journals/tits/FangYQXY22} is adopted. For the VD task, we implement a visual decision work (VD-OIA)\footnote{\url{https://twizwei.github.io/bddoia_project/}} by adopting the same training and testing frames on action decision from VD-OIA, while differently, we load our model (\textbf{C}) on the training data (\textbf{R}) to obtain a difference map (\textbf{R}-\textbf{C}(\textbf{S})) for better important object learning (named as \underline{VD-OIA*}). For the TAA task, we use the same AP, AUC, and TTA$_{0.5}$ of \cite{CognitiveTAA} for an accident frame and temporal occurrence determination, and the Action mF1, and Action F1-all metrics in VD-OIA for VD task. From Tab.~\ref{tab1} and Fig.~\ref{fig20}, our model achieves better performance than the baselines. Notably, Cog-TAA (Ours) trained on accident-free videos achieves better AP and TTA$_{0.5}$ values than Cog-TAA trained on manually-labeled accident videos. 
    \begin{figure*}[htpb]
  \centering
\includegraphics[width=\linewidth]{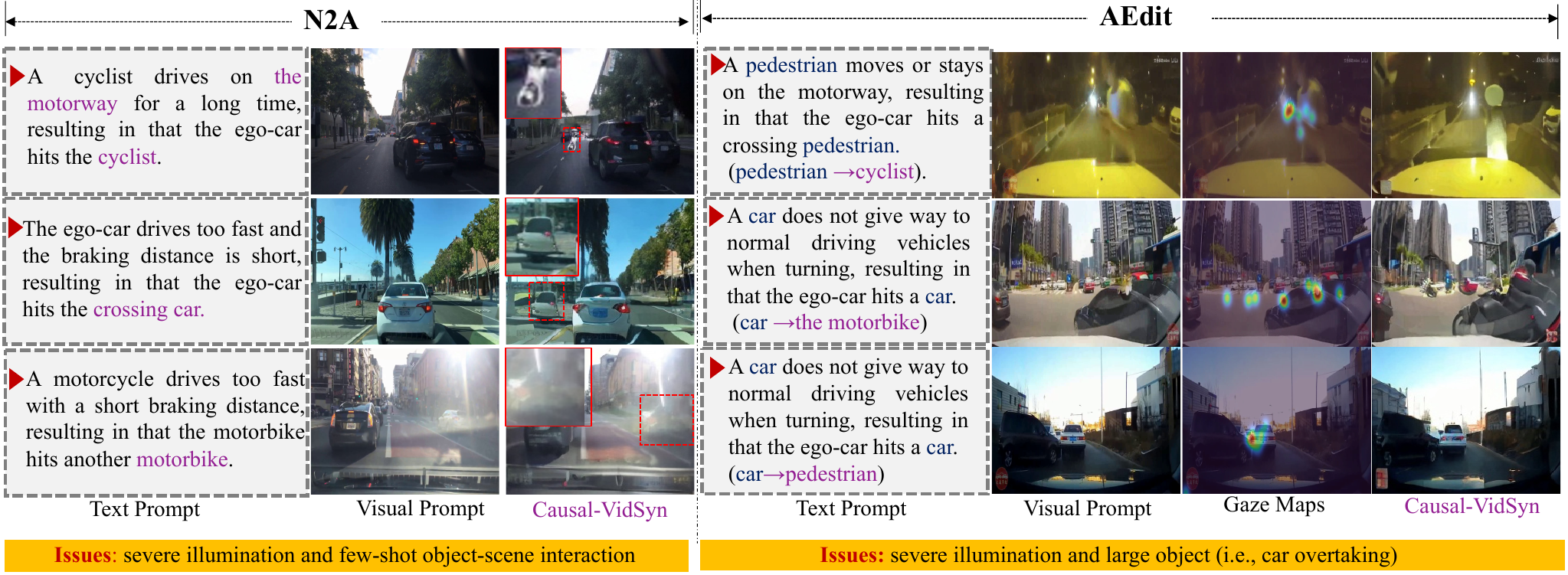}
   \caption{\small{Some failure cases in N2A and AEdit tasks.}}
   \label{fig16}
\end{figure*}

\section{Failure Case Analysis}
In addition, we also show the limitations of our \modelname~by analyzing some failure cases, as shown in Fig.~\ref{fig16}.
In this analysis, we take several samples in N2A and AEdit tasks because of the demand for causal-sensitivity checking. For the failure cases in the N2A task, we can observe that the expected video content change does not appear because of the severe illumination (strong or dark light conditions) and rare object-scene interaction (\emph{i.e., a cyclist drives on the motorway for a long time}). These failure cases inspire two possible insights: 1) Light or weather conditions can be further involved in the text prompt and model designs in future research; 2) The appropriate selection of text prompt in video diffusion is important, especially in the mixed traffic scenes. The text and visual frame prompts need to be well-paired for realistic video diffusion. Maybe, the dense video captioning approach (\emph{e.g.}, Pllava\footnote{\url{https://pllava.github.io/}}) can be introduced with a collision phrase (``hitting'') injection.

Additionally, for the AEdit task, besides the severe illumination issue, the large-scale objects (occupying large regions) are hard to be edited (\emph{e.g.}, the cars with ``overtaking'' behavior). For these situations, some object detectors can be further adopted, and the causal-entity editing in the egocentric accident video synthesis needs to consider some object-level insights. However, involving more conditions may restrict the flexibility of the causal token selection and grounding process. Therefore, this remains an open problem and needs to be considered in the future. 

From the extensive visualization and ablation analysis, the superiority of our \modelname~is evidently verified.

\section{Ethics Statement}
The misuse including the creation of deceptive accident content for evidence collection may have negative societal impacts, and we advocate positive use for deep accident understanding, such as accident anticipation. In addition, we claim that all authors have solid contributions to this work.

\end{document}